\pdfoutput=1
\documentclass[11pt]{article}
\usepackage{authblk}
\usepackage{EMNLP2022}
\usepackage{times}
\usepackage{latexsym}
\usepackage[T1]{fontenc}
\usepackage[utf8]{inputenc}
\usepackage{microtype}
\usepackage{inconsolata}

\usepackage{fullpage}
\usepackage{enumitem}
\usepackage{pgfplots}
\usepackage{algorithm}
\usepackage{algpseudocode}
\usepackage{graphicx}
\usepackage{placeins}
\usepackage{tabularx}
\usepackage{makecell}
\usepackage{booktabs}       
\usepackage{array}          
\usepackage{colortbl}
\newcolumntype{P}[1]{>{\centering\arraybackslash}p{#1}}
\newcolumntype{M}[1]{>{\centering\arraybackslash}m{#1}}
\newcommand{\greyrule}{\arrayrulecolor{black!30}\midrule\arrayrulecolor{black}}
\usepackage{subcaption}
\usepackage{amssymb}
\usepackage{amsmath}
\usepackage{mathtools}
\usepackage{hyperref}

\newcommand{\vect}[1]{\mathbf{#1}}
\newcommand{\wm}{\vect{w}_M}
\newcommand{\ws}{\vect{w}^*}
\newcommand{\w}{\mathbf{w}}
\newcommand{\hess}{\vect{H_{\mathcal{L}}}}
\newcommand{\pr}[1]{\left(#1\right)}
\newcommand{\dw}{\delta \w}
\newcommand{\dL}{\delta \mathcal{L}}
\newcommand{\eF}{\widehat{\vect{F}}}

\newcommand{\e}{\vect{e}}
\newcommand{\E}{\vect{E}}
\newcommand{\subQ}{\textrm{Q}}

\def\1{\bm{1}}

\newcommand{\comment}[1]{}

\newcommand{\bert}{$\textrm{BERT}_{\textrm{BASE}}\,$}
\newcommand{\bertL}{$\textrm{BERT}_{\textrm{LARGE}}\,$}
\pgfplotsset{width=8cm,compat=1.9}
\graphicspath{ {./media/} }

\title{The Optimal BERT Surgeon: Scalable and Accurate Second-Order Pruning for Large Language Models}
\begin{document}

\author[1]{Eldar Kurtic\thanks{~~~Corresponding author: eldar.kurtic@ist.ac.at.}~~}
\author[2,3]{Daniel Campos}
\author[2]{Tuan Nguyen}
\author[1]{Elias Frantar}
\author[2]{Mark Kurtz}
\author[2]{\\Benjamin Fineran}
\author[2]{Michael Goin}
\author[1,2]{Dan Alistarh}
\affil[1]{Institute of Science and Technology Austria}
\affil[2]{Neural Magic Inc.}
\affil[3]{Department of Computer Science, University of Illinois Urbana-Champaign}

\maketitle
\begin{abstract}
In this paper, we consider the problem of sparsifying BERT models, which are a key building block for natural language processing, in order to reduce their storage and computational cost. 
We introduce the \emph{Optimal BERT Surgeon} (oBERT), 
an efficient and accurate pruning method based on approximate second-order information, which we show to yield state-of-the-art results for compression in both stages of language tasks: pre-training and fine-tuning. Specifically, oBERT extends existing work on second-order pruning by allowing for pruning blocks of weights, and is the first such method that is applicable at BERT scale. 
Second, we investigate \emph{compounding} compression approaches to obtain highly compressed but accurate models for deployment on edge devices. These models significantly push boundaries of the current state-of-the-art sparse BERT models with respect to all metrics: model size, inference speed and task accuracy. For example, relative to the dense \bert, we obtain 10x model size compression with \hbox{< 1\%} accuracy drop, 10x CPU-inference speedup with \hbox{< 2\%} accuracy drop, and 29x CPU-inference speedup with \hbox{< 7.5\%} accuracy drop. Our code, fully integrated with Transformers and SparseML, is available at
\url{https://github.com/neuralmagic/sparseml/tree/main/research/optimal\_BERT\_surgeon\_oBERT}.\looseness=-1

\end{abstract}
\section{Introduction} 

\begin{figure}[t]
    \centering
    \includegraphics[scale=0.5]{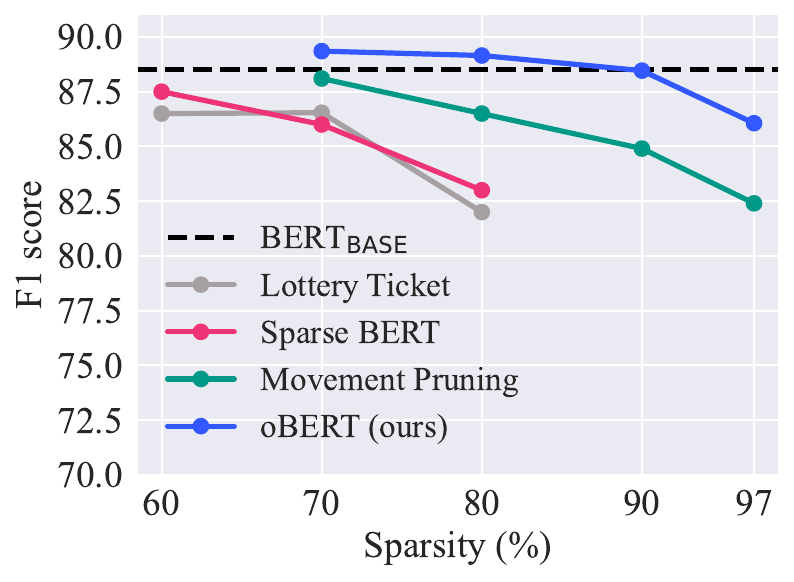}
    \vspace{-0.1in}
    \caption{Performance overview relative to state-of-the-art unstructured downstream pruning methods \citet{chen2020lottery}, \citet{Xu2021RethinkingNP}, \citet{Sanh2020MovementPA}, in this order, of the \bert model on the SQuADv1.1 task.}
    \label{fig:squad_F1}
    \vspace{-0.3in}
\end{figure}

Pre-trained Transformer models~\cite{Vaswani2017AttentionIA, Devlin2019BERTPO} provide robust language representations which can be specialized on various tasks. Given their massive  growth~\cite{Radford2019LanguageMA, MTNLG}, techniques for reducing their computational overheads have become popular. One classic technique is Knowledge Distillation (KD)~\cite{Hinton2015DistillingTK}, which transfers knowledge from a larger teacher to a smaller student model. Other work has leveraged lower-precision representations to produce quantized models. An orthogonal approach, which is our primary focus, has been to apply unstructured and block pruning, i.e. removing individual weights, to produce compressed but accurate language models. \mbox{Figure~\ref{fig:squad_F1}} provides a comparative overview of state-of-the-art results for unstructured pruning.\looseness=-1

In this paper, we introduce a method for improved unstructured and semi-structured (block) pruning, by leveraging the second-order approach pioneered by the Optimal Brain Surgeon framework~\cite{LeCun1989OptimalBD, hassibi1993second}, which we scale for the first time to LLMs. Further, we put our results in the context of a compound compression approach, which combines several compression techniques to obtain sparse models which we execute on a sparsity-aware CPU-based runtime~\cite{deepsparse}, showing order-of-magnitude speedups at low accuracy loss.

\noindent In summary, our contributions are as follows:
\setlist{nolistsep}
\begin{itemize}[noitemsep]
\setlength{\itemindent}{0.5em}
    \item We perform a thorough exploration of weight pruning approaches applied to LLMs, including lottery-ticket, movement pruning, magnitude and second-order pruning. 
    
    \item We introduce a general second-order pruning method called \emph{Optimal BERT Surgeon} (oBERT), which supports unstructured and block pruning, and is the first second-order method to be both highly-accurate and scalable to the dimensionality of BERT models. 
    
    \item We illustrate the benefits of oBERT by significantly improving upon existing state-of-the-art pruning methods, in both stages of language tasks: pre-training and fine-tuning. For illustration, when pruning \bert, oBERT outperforms Movement Pruning (MvP), the most accurate prior approach, by more than 2\% absolute F1 score at the same sparsity, and can match the accuracy of MvP models with 3x fewer parameters. 
    
    \item We investigate the applicability of this pruning method in a framework which \emph{compounds} popular compression approaches for LLMs, i.e. applying pruning in combination with layer dropping and/or quantization. In this context, we show that our resulting sparse models provide order-of-magnitude improvements compared to other compound compressed models, and that they can be easily deployed for CPU inference.
\end{itemize}
 
\section{Background and Related Work}

\noindent\textbf{Transformer LLMs} are usually built using multiple transformer layers with self-attention~\cite{Vaswani2017AttentionIA}. Each transformer has a variation of two sub-components: multi head attention (MHA) and fully connected feed forward network (FFN). 
Given the massive size of well-performing models, there has been growing interest in LLM compression. They have been shown to be fragile as minor perturbations can lead to model collapse~\cite{DBLP:journals/corr/abs-2105-06990}. Pruning schemes are motivated by weight saliency metrics which represent the loss in accuracy due to pruning. It is common to prune in iterative steps, each of which removes weights until a desired sparsity level is reached. Now, we briefly overview existing approaches.\looseness=-1

\noindent\textbf{Structured pruning} for LLMs focuses on reducing the number of layers and/or attention heads, and requires structural understanding of the model. 
\citet{Michel2019AreSH} and \citet{Voita2019AnalyzingMS} demonstrated that for some tasks nearly 40\% of attention heads can be removed without major impact on accuracy. 
Other work has focused on removing layers~\cite{Sridhar2020UndividedAA}, and on the order in which they are removed~\cite{DBLP:journals/corr/abs-2004-03844}. In some of our experiments, we apply standard ``direct'' layer dropping in conjunction with pruning.

\noindent\textbf{Semi-structured pruning} is an intermediate approach, by which smaller groups, e.g. rectangular sets of weights~\cite{lagunas21block}, are set to zero. This approach has recently gained in popularity thanks to efficient computational support. We extend the second-order pruning formulation to such groupings, and show results for a specific grouping supported by a CPU-inference engine.

\noindent\textbf{Unstructured pruning} removes individual weights by setting them to zero.
Gradual Magnitude Pruning (GMP) is a classic approach, which makes use of weight magnitudes as a saliency metric for pruning~\cite{Han2015ADN, Gale2019TheSO}.\looseness=-1

\noindent\textbf{First-order pruning} methods use a gradient based formulation of the saliency metric. A popular method is \textbf{Movement Pruning (MvP)}~\cite{Sanh2020MovementPA}, specifically designed for pruning in the fine-tuning stage. Intuitively, it removes weights that are moving towards zero. The resulting models were the first to achieve high sparsity with tolerable accuracy loss. Methods such as PLATON~\cite{zhang2022platon} attempt to capture the uncertainty of weights importance scores by upper confidence bound estimation. Prior to our work, MvP and PLATON approaches set state-of-the-art results for unstructured pruning.\looseness=-1

\noindent\textbf{Second-order pruning} methods~\cite{LeCun1989OptimalBD, hassibi1993second, Singh2020WoodFisherES, Frantar2021EfficientMA} were developed in the context of image classification, and leverage complex approximations of the loss curvature. 
However, second-order pruning methods require an approximation of the inverse Hessian, which is expensive to store and compute with for LLM parameter counts. The approach we propose is similar to  WoodFisher/M-FAC methods~\cite{Singh2020WoodFisherES, Frantar2021EfficientMA}, but is the first to work accurately at LLM scale. Specifically, the WoodFisher approach is infeasible at BERT scale, as it requires storing gradients for inverse Fisher calculation in memory at the point of pruning. 
The M-FAC approach scales, but we show that its parametrization yields worse pruning results (Appendix Figure~\ref{fig:ablation_B}). 
This is because M-FAC performs full-matrix (non-blocked) inversion by default, which is inherently noisy. 
In addition, we extend the theoretical OBS approach to semi-structured (block) compression. We also show that our method can be applied during LLM pre-training and fine-tuning, yielding state-of-the-art results in both regimes.

\noindent\textbf{Knowledge Distillation}~\cite{Hinton2015DistillingTK} trains a smaller student model against outputs of a larger teacher model by adding a loss component which minimizes the KL-divergence between the two output distributions, which is the approach we adopt in our setup too. A hardness parameter is used to control the mixture of regular and distillation loss, and a temperature parameter to control softness of the distribution. Contrary to this, approaches like DistilBERT~\cite{Sanh2019DistilBERTAD}, TinyBERT~\cite{Jiao2020TinyBERTDB}, MobileBERT~\cite{Sun2020MobileBERTAC}, and MiniLM~\cite{Wang2020MiniLMDS} utilize more complex distillation schemes, based on transferring knowledge from intermediate model's representations. Our sparse models provide order-of-magnitude improvements upon some of these methods. 

\noindent\textbf{Quantization} represents weights and activations in lower precision~\cite{Courbariaux2016BinarizedNN}, and was used to obtain models such as Q8BERT~\cite{Zafrir2019Q8BERTQ8} and TernaryBERT~\cite{zhang2020ternarybert}.

\citet{shen2020q} uses information about the Hessian spectrum to choose quantization bit-widths, whereas \citet{yu2022hessian} uses an approximation of the Hessian trace for structured pruning.
These Hessian-based approaches are different from the one we propose, as we use completely different inverse-Hessian approximations to guide pruning decisions. 
The focus of our work is on \emph{weight pruning}, and on computational speedups achievable on commodity CPUs. 
As such, the methods we investigate are orthogonal to quantization. 
Moreover, it is impossible to directly compare to low-bitwidth quantized models as most inference frameworks do not support such custom formats. Therefore, we will only make use of the standard Quantization-Aware Training (QAT) to 8-bit weights, which is well-supported on Intel CPUs, and showcase the resulting speedups in conjunction with layer dropping and weight pruning. 

\noindent\textbf{Downstream compression} methods attempt to compress directly while fine-tuning on a specific task. MvP method is specially designed for this setup. \textbf{Upstream compression} methods compress during the pre-training phase, reducing the need for task-specific pruning. \citet{chen2020lottery} examined the ``Lottery Ticket'' strategies~\cite{Frankle2019TheLT} which, as we illustrate later, incur huge accuracy loss even at moderate sparsities. Recent work ``Prune Once for All'' (Prune OFA) by~\citet{zafrir2021prune} showed that well-tuned magnitude pruning can be competitive with downstream methods like MvP.\looseness=-1

We first examine the performance of prior pruning methods, notably MvP, Prune OFA, and Lottery Tickets, relative to the new second-order oBERT method. The approach we propose consistently improves upon all these prior methods, both in pre-training (upstream) and fine-tuning (downstream) stages, and can be compounded with other compression techniques to obtain models that are smaller, faster and more accurate than models like DistilBERT, TinyBERT, and block MvP. 

Additional approaches for efficient inference of LLMs exist, like token-pruning and early-exiting. These approaches are \emph{orthogonal} to ours; therefore we discuss them in Appendix~\ref{app:additional_comparisons}.

\section{The Optimal BERT Surgeon (oBERT)}
\label{sec:obs}

\subsection{Generalized Second-Order Block Pruning}

The pruning problem starts from a well-optimized dense model \( \ws \in \mathbb{R}^d \), and aims to find a sparse version of $\ws$, where many of the weights are set to zero, 
and the remaining weights may be updated accordingly in order to preserve the loss. 
It is common for this process to occur gradually, i.e. by progressively removing the weights. 
A classic approach~\cite{LeCun1989OptimalBD, hassibi1993second}  for ``optimal'' pruning of weights from $\ws$ at a step is to expand the loss function $\mathcal{L}$ locally around $\ws$ with respect to a sparse 0/1 weight mask $\vect{M}$. If we denote by $\w_M = (\vect{M} \odot \ws)$, the model resulting from the Hadamard (element-wise) product between \( \vect{M} \in \{0,1\}^d \) and $\ws$, we can use the Taylor expansion at $\w_M$ to obtain: 
\begin{eqnarray*}
    \mathcal{L}(\wm) \simeq \mathcal{L}(\ws) + (\wm - \ws)^\top \nabla \mathcal{L}(\ws) \\ + \frac{1}{2} (\wm - \ws)^\top \hess (\ws) (\wm - \ws). 
\end{eqnarray*}

\noindent Given that $\ws$ is well-optimized, it is reasonable in practice to assume that \( \nabla \mathcal{L}(\ws) \approx \vect{0} \).
Then, the change in loss incurred by pruning a subset of weights can be expressed as
\begin{equation}
\label{eq:delta-loss}
    \dL (\dw) \simeq \frac{1}{2} \dw^\top \hess (\ws) \dw 
\end{equation}

\noindent where \( \dL (\dw) \coloneqq \mathcal{L}(\wm) - \mathcal{L}(\ws) \) and \( \dw \coloneqq \wm - \ws \). 
A popular way of approximating the Hessian at $\ws$ is via a dampened  empirical Fisher information matrix~\cite{hassibi1993second}:
\begin{equation}
\label{eq:emp-fisher}
\noindent
    \hess(\w)\hspace*{-.25em} \simeq \hspace*{-.25em} \eF (\w) \hspace*{-.25em}=\hspace*{-.25em} \lambda \vect{I}_d + \frac{1}{m} \sum_{i=1}^{m} \nabla \mathcal{L}_i(\w) \nabla \mathcal{L}^\top_i(\w),
\end{equation}
\noindent where \( \lambda \geq 0 \) is a small dampening constant, \( \vect{I}_d \in \mathbb{R}^{d \times d} \) identity matrix and \( m \) is the number of gradient outer products used to approximate the Hessian. Given the positive-definiteness of \eqref{eq:emp-fisher}, the quadratic form \eqref{eq:delta-loss} is always nonnegative which is why we will refer to \( \dL(\dw) \) as a \textit{loss increase} incurred by pruning. 

Returning to our pruning problem, assume we wish to identify a block of weights $Q$ of a given shape whose removal by zero-masking would incur minimum increase in loss. This leads to the following constrained optimization problem: 
\begin{equation}
\label{eq:opt}
\begin{split}
    &\min_{\dw}\quad \frac{1}{2} \dw^\top \eF (\ws) \dw \\&\phantom{x}\mathrm{s.t.}\quad \e_k^\top \dw + w_k = 0, \quad \forall k \in \textrm{Q} 
\end{split}
\end{equation}

\noindent where \( \e_k \in \mathbb{R}^d \) stands for the \( k \)-th canonical basis vector. Here, we will provide a generalized solution, which applies to general $Q$. 
First, for convenience, we express the system of \( |\textrm{Q}| \) equality constraints in matrix-equation form as
\(    \E_\subQ \dw + \E_\subQ\ws = \vect{0}, \)
 where \( \E_\subQ \in \mathbb{R}^{|\subQ| \times d} \) is a matrix composed of the corresponding canonical basis vectors \( \e_k\, (\forall k \in \subQ) \) arranged in rows. This optimization problem can be solved with the method of Lagrange multipliers. Specifically, we wish to find stationary points of the Lagrangian \( L(\dw, \boldsymbol{\lambda}) \), where \( \boldsymbol{\lambda} \in \mathbb{R}^{|\subQ|} \) denotes a vector of Lagrange multipliers. Solving the system of equations 
\(
    \frac{\partial L(\dw, \boldsymbol{\lambda})}{\partial \dw} = \vect{0} \) and 
\(    \frac{\partial L(\dw, \boldsymbol{\lambda})}{\partial \boldsymbol{\lambda}} = \vect{0} \)
yields the following optimal weight update:
\begin{equation*}
\label{eq:update}
    \dw^* = -\eF^{-1}\hspace*{-.1em}(\ws)\E_\subQ^\top\hspace*{-.2em}\pr{\E_\subQ \eF^{-1}\hspace*{-.1em}(\ws) \E_\subQ^\top}^{-1}\hspace*{-.4em}\E_\subQ \ws 
\end{equation*}
which prunes a set of weights $Q$ and updates the remaining weights to preserve the loss.
Now, the corresponding loss increase incurred by the optimal weight update \( \dw^* \) can be expressed as the saliency score of weights $Q$, which we denote by:
\begin{equation*}
\label{eq:saliency}
    \rho_\subQ = \frac{1}{2} \pr{\E_\subQ \w^*}^\top \pr{\E_\subQ \eF^{-1}(\ws)\E_\subQ^\top}^{-1} \E_\subQ \w^*.
\end{equation*}

We use this saliency/importance score to rank groups of weights for pruning. As a sanity check, if we prune a single weight $w_j$ at a time, our derivations will yield the standard formulas of~\cite{hassibi1993second}. The full version of~\citet{Singh2020WoodFisherES} provided a slightly less general derivation for the blocked case, under additional assumptions.  

\subsection{An Efficient Implementation}

Directly implementing the previously described approach for LLMs, where number of weights $\w \in \mathbb{R}^d$ is huge, is infeasible. In particular, this is due to the dependence on the inverse of the empirical Fisher information matrix $\eF^{-1}(\w) \in \mathbb{R}^{d \times d}$, appearing in formulations of the saliency score and of the optimal weight update. 
We now describe how to circumvent these issues. 

\subsubsection{Pruning the optimal set of weights}

Assume a gradual pruning setup, in which at each pruning step we wish to prune a model to a target sparsity $s \in (0, 1]$, effectively zeroing out $s \times d$ weights, in groups of size $|Q|$. Typically $s \times d \gg |Q|$, meaning that we want to remove multiple groups at the same time. Finding the optimal set of $\frac{s \times d}{|Q|}$ groups is an intractable combinatorial problem, due to all possible correlations between them, given by the binomial coefficient $\binom{n}{k}$, where $n = \frac{d}{|Q|}$ and $k = \frac{s \times d}{|Q|}$. 
This problem can be alleviated by ignoring correlations between different groups of weights $Q$, and solving only for correlations between the weights within the same group. 
In practice, this boils down to evaluating the saliency score $\rho_\subQ$ for each group $Q$, and pruning the $\frac{s \times d}{|Q|}$ groups with the lowest score. 
As pruning many weights in the same step can make the Taylor approximation of the loss function less accurate, one can consider pruning with multiple smaller sub-steps with recomputations of the Hessian approximation in between (without intermediate fine-tuning). 
While this can further improve the quality of the pruning step~\cite{Frantar2021EfficientMA}, we do not implement this additional optimization since the competing methods do not utilize recomputations. 

\subsubsection{Inverse empirical Fisher computation}

The key space and time complexity cost of the above procedure is computing products with the inverse empirical Fisher. 
A direct approach would be to perform a block-wise diagonal approximation of this matrix (which we detail next), and perform direct block inversion.
However, we found experimentally that this approach is too expensive in terms of time, and quite numerically-sensitive. 
As an alternative, we rely on the fact that the matrix we wish to invert is a sum of rank-1 matrices, and employ the Woodbury/Sherman-Morrison (WSM) inversion formula. 
Specifically, given a sum $(\vect{A} + \vect{u}\vect{v}^\top)$ of an invertible matrix $\vect{A}$ and an outer product of vectors $\vect{u}$ and $\vect{v}$ with compatible dimensions, the inverse $(\vect{A} + \vect{u}\vect{v}^\top)^{-1}$ can be exactly calculated as $\vect{A}^{-1} - \frac{\vect{A}^{-1}\vect{u}\vect{v}^\top\vect{A}^{-1}}{1+\vect{v}^\top\vect{A}^{-1}\vect{u}}$. Placing the expression of the empirical Fisher in the WSM formula, we obtain the following recursive formulation, where $m$ is the number of gradients employed in the approximation:
\begin{eqnarray*}
    & \eF^{-1}(\w) = \eF^{-1}_m(\w) = \\ & \pr{\eF_{m-1}(\w) + \frac{1}{m} \nabla \mathcal{L}_m(\w) \nabla \mathcal{L}_m^\top(\w)}^{-1}. &
\end{eqnarray*}
\noindent Unrolling the recursion with $\eF^{-1}_0(\w) = \frac{1}{\lambda} \vect{I}_d$, we can obtain an iterative formula to exactly calculate the inverse of the empirical Fisher matrix as
\begin{eqnarray*}
\label{eq:iterative}
    & \eF^{-1}(\w) =  \eF^{-1}_m(\w) = \\ & \frac{1}{\lambda} \vect{I}_d - \sum_{i=1}^{m} \frac{\pr{\eF^{-1}_{i-1}(\w) \nabla \mathcal{L}_i(\w)}\pr{\eF^{-1}_{i-1}(\w) \nabla \mathcal{L}_i(\w)}^\top}{m + \nabla \mathcal{L}_i^\top(\w) \eF^{-1}_{i-1}(\w) \nabla \mathcal{L}_i(\w)}. &
\end{eqnarray*}
The iterative formulation enjoys a number of computational advantages over the direct implementation. The most notable ones are 1) avoiding explicit calls to the expensive and dampening-sensitive matrix inversions, and 2) allowing successive updates of the inverse as new gradients are computed, never needing to store all $m$ gradients of size $d$ and thus significantly reducing memory requirements. 

\subsection{Memory and run-time complexity}

Computing and storing the inverse empirical Fisher $\eF^{-1}(\w) \in \mathbb{R}^{d \times d}$ is prohibitively expensive for modern LLMs, which have hundreds of millions of parameters, due to the quadratic complexity on the number of weights $d$. However, \citet{Singh2020WoodFisherES} have shown that a diagonal block-wise approximation of the empirical Fisher matrix can be very accurate for pruning of convolutional neural networks.  We adapt the same approach here, in the context of LLMs. Thus, for blocks of width $B$ along the main diagonal, memory requirements for the computation of the inverse Fisher matrix are reduced from the quadratic $\mathcal{O}(d^2)$ to a linear $\mathcal{O}(Bd)$ dependence on the number of weights $d$. At the same time, run-time complexity relaxes from $\mathcal{O}(md^2)$ to $\mathcal{O}(mBd)$. 
As we will show, this computation can be efficiently and accurately performed for moderate values of $m$ and $B$. 

{Another alternative we investigated was the matrix-free approach of~\citet{Frantar2021EfficientMA}, which does not require a block-wise approximation and has complexity $\Theta(dm)$. However, our investigation showed that this approach required high values of $m$ to be accurate (Appendix Figure~\ref{fig:ablation_B}), which leads to excessive memory cost in the case of BERT models.} 

\subsection{Efficient and scalable implementation}

On the practical side, we have identified general hyper-parameters $B = 50$ for the block size, and $m = 1024$ for the number of gradients which produce state-of-the-art results for all analyzed BERT models (for more details please see Appendix~\ref{app:obs-hyperparams}), while still being able to fit on the 24GB RTX 3090 GPU. We reflect upon the computational costs in more detail in Appendix~\ref{app:computational_costs}. 
Moreover, for these parameter values, the block-wise approximation of $\eF^{-1}(\w)$ can be implemented very efficiently on modern accelerators. Specifically, we take advantage of the fact that such hardware favors batched matrix operations, and that the blocks of size $B \times B$ in $\eF^{-1}(\w)$ are independent. With $N_B = \frac{d}{B}$ we refer to the total number of blocks, {i.e.} the batch-dimension. 
\noindent The procedure works as follows. First, we compute batched matrix-vector products $\eF^{-1}_{i-1}(\w) \nabla \mathcal{L}_i(\w) \in \mathbb{R}^{N_B \times B}$ and scalar denominators
$
m + \nabla \mathcal{L}^\top_i(\w) \eF^{-1}_{i-1}(\w) \nabla \mathcal{L}_i(\w) \in \mathbb{R}^{N_B}.
$
\noindent Then, we update the inverse Fisher for each block by computing the scalar-scaled outer products 
$
\pr{\eF^{-1}_{i-1}(\w) \nabla \mathcal{L}_i(\w)} \pr{\eF^{-1}_{i-1}(\w) \nabla \mathcal{L}_i(\w)}^\top
$
of shape $\mathbb{R}^{N_B \times B \times B}$.

\section{Experimental Validation}
\label{sec:experiments}
To ease reproducibility, we conduct our experiments in modified versions of the popular open-source libraries: Transformers~\cite{wolf-etal-2020-transformers}, and SparseML~\cite{pmlr-v119-kurtz20a}. All of our experiments are using publicly available datasets via~\citet{hf-datasets} and focus on the \bert model~\cite{Devlin2019BERTPO}, one of the most commonly used LLMs, composed of 12 transformer layers with 110M parameters. Following community standards, we prune encoder's weights (85M) and report sparsities relative to this number. All of our models, compression recipes and the full implementation will be made public.

\subsection{Downstream Unstructured Pruning}
\label{sec:downstream}

We first revisit the accuracy-compression trade-off for pruning on downstream tasks. 

\noindent\textbf{Goals and setup.} 
We compare existing approaches, notably Movement Pruning (MvP)~\cite{Sanh2020MovementPA} and Lottery Ticket (LT-BERT)~\cite{chen2020lottery}, against the gradual unstructured oBERT method, introduced in Section~\ref{sec:obs}. Our experiments evaluate performance on a variety of downstream (English) tasks commonly used to evaluate model compression: question answering SQuAD v1.1 \cite{Rajpurkar2016SQuAD1Q}, sentence classification Quora Duplicate Query Dataset QQP \cite{shankar2017first}, and natural language inference MNLI \cite{N18-1101}.

\noindent\textbf{Comparison with MvP.} For a fair comparison with MvP, we consider the 10-epoch gradual pruning setup used to obtain the best results by~\citet{Sanh2020MovementPA}. Specifically, we start from the \bert model and perform 2 epochs of fine-tuning, followed by 6 epochs of pruning, and 2 further epochs of fine-tuning of the compressed model. We impose a global sparsity distribution over all layers, prune with oBERT two times per epoch, and use KD from the fine-tuned \bert teacher. For oBERT pruning we use $m=1024$ gradients, block size $B=50$, and dampening $\lambda = 10^{-7}$ to approximate the inverse Hessian matrix. In all of our runs, the first pruning step prunes 70\% of weights and then follows the cubic interpolation~\cite{Zhu2018ToPO} to the target sparsity. This large first pruning step gives more time to recover from the later pruning steps, which impose higher sparsities. All  hyper-parameters are described in detail in Appendix~\ref{app:hyperparams-DownstreamPruning}, and the results are given in Table \ref{tab:10_30_gradual} (in the \textit{10 Epochs} section). 

We observe that Optimal BERT Surgeon outperforms Movement Pruning by a significant margin, more than 2 points of F1 score at the same sparsity. Remarkably, the model pruned with oBERT to 97\% sparsity has similar accuracy to MvP-pruned model at 90\% sparsity, which has roughly 3x more weights. This reinforces the effectiveness of second-order information for pruning.

\noindent\textbf{Extended pruning and fine-tuning.} 
Next, we examine effects of extending the gradual schedule to 30 epochs, matching the setup used for LT-BERT~\cite{chen2020lottery}. The only difference compared to our 10 epoch setup is that we now prune with oBERT every four epochs, and rewind learning rate after each pruning step. The extended setup leaves more time to recover from pruning, which reflects in the improved results in Table~\ref{tab:10_30_gradual} (\textit{30 Epochs} section). We report the mean over three runs. For additional evaluation metrics and standard deviations please see Tables \ref{tab:10_30_gradual_v2} and \ref{tab:10_30_gradual_stdev} in the Appendix. The results show a clear accuracy difference between oBERT and LT-BERT, especially at high sparsities. This difference is justified since the LT based approach attempts to mainly transfer \emph{network connectivity}, whereas the oBERT can also benefit from the weight values. Finally, we examined the impact of extended setup with Soft MvP on SQuAD, targeting 90\% sparsity (not shown in the Table), leading to an (F1, EM) combination of $(87.42, 79.83)$ for MvP. The F1 gap in favor of oBERT is lower than at 10 epochs, suggesting that extended finetuning helps all methods; yet, it is far from negligible.

\begin{table}
\setlength{\tabcolsep}{4pt}
    \caption{Downstream tasks dev-set performance of pruned \bert models. ($^*$ approximate results as the exact numbers are not available.)}
    \label{tab:10_30_gradual}
    \centering
    \scalebox{0.93}{
    \small{
    \begin{tabular}{ccc|cc|cc}
    \toprule 
    Task & \makecell{BERT\\{\scriptsize{BASE}}} & Spars. & \makecell{Soft\\MvP} & \makecell{oBERT\\(ours)} & \makecell{LT-\\BERT} & \makecell{oBERT\\(ours)}\\
    \midrule
      & Epochs && \multicolumn{2}{c|}{10 Epochs} & \multicolumn{2}{c}{30 Epochs}\\  
    \midrule
    \makecell{SQuAD\\F1} & 88.54 & \makecell{80\% \\ 90\% \\ 97\%} &  \makecell{- \\ 84.90 \\ 82.30} &  \makecell{- \\ \textbf{87.98} \\ \textbf{84.65}} & \makecell{86.54\phantom{$^*$} \\ 68.00$^*$ \\ -\phantom{$^*$}} & \makecell{\textbf{89.04} \\ \textbf{88.31} \\ \textbf{85.98}}\\
    \midrule
    \makecell{MNLI\\m-acc} & 84.54 & \makecell{80\% \\ 90\% \\ 97\%}  &  \makecell{- \\ 81.20 \\ 79.50} & \makecell{- \\ \textbf{83.20} \\ \textbf{81.00}} & \makecell{82.60\phantom{$^*$} \\ 75.00$^*$ \\ -\phantom{$^*$}} &  \makecell{\textbf{84.32} \\ \textbf{83.79} \\ \textbf{81.77}}\\
    \midrule
    \makecell{QQP\\Acc} & 91.06 &  \makecell{80\% \\ 90\% \\ 97\%} & \makecell{- \\ 90.20 \\ 89.10} & \makecell{- \\  \textbf{90.89} \\ \textbf{90.23}} &  \makecell{90.30\phantom{$^*$} \\ 90.00\phantom{$^*$} \\ -\phantom{$^*$}} & \makecell{\textbf{91.57} \\ \textbf{91.35} \\ \textbf{90.87}}\\
    \bottomrule
    \end{tabular}
    }
    }
    \vspace{-1.2em}
\end{table}

\subsection{Upstream Unstructured Pruning}
\label{sec:upstream}

An appealing alternative to downstream pruning is to compress models upstream, on the semi-supervised pre-training task~\cite{zafrir2021prune}. Given the upstream pruned model, computational  requirements for obtaining downstream fine-tuned models are significantly reduced, as only fine-tuning of the remaining weights is necessary.

\noindent\textbf{Goals and setup.} To compare with existing approaches, notably Prune OFA~\cite{zafrir2021prune} and LT-BERT~\cite{chen2020lottery}, we gradually prune with oBERT directly at upstream datasets, BookCorpus and English Wikipedia, and then fine-tune the remaining unpruned weights on the subset of GLUE tasks.

\noindent\textbf{Teacher preparation.} Following~\citet{Liu2019RoBERTaAR}, we start with the HuggingFace \bert uncased model, and fine-tune it for additional 10 epochs only on the masked language modeling task.

\noindent\textbf{Pruning at upstream.} Once the distillation teacher is trained, we gradually prune and fine-tune the \bert model for 3 epochs, using KD from the dense teacher. We prune four times per epoch, and rewind learning rate to the initial value after each pruning step. Hyper-parameters for oBERT are the same as for downstream pruning in~\ref{sec:downstream}; a full description can be found in Appendix~\ref{app:hyperparams-UpstreamPruning}.

\noindent\textbf{Sparse-transfer to downstream.} To evaluate the resulting upstream-pruned models, we finetune the unpruned weights on downstream tasks with KD from the fine-tuned \bert model. For a fair comparison with Prune OFA, we fine-tune for 8 epochs. The results in Table~\ref{tab:sparse-transfer} show that sparse models produced by oBERT outperform state-of-the-art methods by significant margins. We report the mean over four runs. For additional evaluation metrics and standard deviations please see Appendix Tables \ref{tab:upstream-additional} and \ref{tab:sparse-transfer-deviations}. It is worth emphasizing that in contrast to Prune OFA, which performed extensive hyper-parameter tuning for sparse-transfer, our recipe is simple and general across downstream tasks: 8 epochs of fine-tuning with linearly decaying learning rate. 
This suggests that sparse pre-trained models found by oBERT constitute a strong starting point for sparse transfer learning, which can be further improved by task-specific hyper-parameter tuning.

\begin{table}[tb!]
\setlength{\tabcolsep}{5.5pt}
    \caption{Sparse-transfer dev-set performance of upstream-pruned \bert models. ($^*$ approximate results as the exact numbers are not available.)}
    \label{tab:sparse-transfer}
    \centering
    {\small 
    \begin{tabular}{ccc|ccc}
    \toprule 
    Task & \makecell{BERT\\{\scriptsize BASE}} & Sparsity & \makecell{LT-\\BERT} & \makecell{Prune\\OFA} & \makecell{oBERT\\(ours)} \\
    \midrule
    \makecell{SQuAD \\ F1} & 88.54 & \makecell{90\% \\ 97\%} & \makecell{68.00$^*$\\-} & \makecell{87.25\\ -} & \makecell{\textbf{88.49}\\ 84.92} \\
    \midrule
    \makecell{MNLI \\ m-acc} & 84.54 & \makecell{90\% \\ 97\%}  & \makecell{75.00$^*$\\-} & \makecell{81.45\\ -} & \makecell{\textbf{83.40} \\ 80.91} \\
    \midrule
    \makecell{QQP \\ Acc} & 91.06 &  \makecell{90\% \\ 97\%} & \makecell{90.00\\-} & \makecell{90.93 \\ -} & \makecell{\textbf{90.99} \\ 90.33} \\
    \midrule
    \makecell{SST-2\\Acc} & 93.01 & 90\% & 85.00$^*$ & 90.88 & \textbf{92.20} \\
    \midrule
    \makecell{QNLI\\Acc} & 91.25 & 90\% & 80.00$^*$ & 89.07 & \textbf{89.97} \\
    \bottomrule
    \end{tabular}
    }
    \vspace{-1.2em}
\end{table}

\subsection{Compound Compression for CPUs}
\label{sec:compound}

To probe the potential practical impact of our approach, we specialize the technique for deployment on CPUs, corresponding to ``edge'' deployments. Specifically, we tailor our sparse models to the DeepSparse~\cite{deepsparse} sparsity-aware runtime, by compounding unstructured pruning with additional compression techniques.  

\noindent\textbf{Direct layer dropping.} The competitive results obtained at high sparsities in sections \ref{sec:downstream} and \ref{sec:upstream} suggest that \bert may be overparameterized for downstream tasks. To improve compression ratio and inference speed, we apply ``direct'' layer dropping: we initially drop all but 3 or 6 of the BERT's 12 layers. We drop layers from our upstream teacher, and, following~\cite{Turc2019WellReadSL}, fine-tune them with KD in the same setup used to prepare the upstream teacher. These 3 and 6 layer models are used as starting points for downstream pruning. More sophisticated layer dropping techniques~\cite{fan2019reducing}, could bring further accuracy gains; we leave this for future work.

\noindent\textbf{Block pruning and QAT.} High-performance inference usually benefits more from (semi) structured sparsity patterns than from the unstructured ones. Hence, we employ the generalized oBERT formulation introduced in the section \ref{sec:obs} and prune weights in the 4-block pattern, meaning that contiguous blocks of 4 weights are either set to zero or kept dense. Both pruning types, unstructured and 4-block, can be leveraged for computational speedups with the DeepSparse runtime, but 4-block pruning coupled with INT8 quantization can provide further performance gains. For quantization, we apply standard quantization-aware training (QAT)~\cite{jacob2018quantization} on top of the 4-block models (see Appendix~\ref{app:hyperparams-DownstreamQuantization} for a full description).

\noindent\textbf{Compounding for deployment.} To determine the impact of different compression schemes, we investigate unstructured and 4-block pruning of the 3, 6, and 12-layer models. For all runs, we use the same set of hyper-parameters from the extended pruning and fine-tuning setup in Section~\ref{sec:downstream}. The results are given in Table~\ref{tab:downstream-block}, where we also report accuracy of the corresponding dense models (0\% sparsity) in the same setup. For additional evaluation metrics, please see Table~\ref{tab:downstream-block-additional}. The results indicate that compression methods can be combined without model collapse, although the accuracy drops do compound. The fact that the layer-dropped models are also highly compressible suggests that structured and fine-grained (unstructured) compression are complementary. We find it remarkable that our 6-layer unstructured oBERT-pruned model is competitive with the 12-layer MvP-pruned model when both are pruned to 90\% sparsity.
\begin{table}
    \centering
    \captionof{table}{F1 score of the 3, 6, and 12-layer models compound-compressed on the SQuADv1.1.}
    \label{tab:downstream-block}
    {\small
    \begin{tabular}{c|c|c|cc}
        \toprule
        Layers & Sparsity & Unstructured & 4-block & +QAT \\
        \midrule
        12 &  \makecell{0\% \\ 80\% \\90\%} & \makecell{89.48 \\ 89.04 \\ 88.31} & \makecell{89.48 \\ 88.57 \\ 87.57} & \makecell{89.06 \\ 87.89 \\ 86.68}\\
        \midrule
        6 & \makecell{0\% \\ 80\% \\90\%} & \makecell{88.32 \\ 88.20 \\ 86.78} & \makecell{88.32 \\ 87.00 \\ 85.34}  & \makecell{87.94 \\ 86.10 \\ 84.59}\\
        \midrule
        3 & \makecell{0\% \\ 80\% \\90\%}  & \makecell{84.66 \\ 84.08 \\ 82.50} & \makecell{84.66 \\ 82.79 \\ 80.69} & \makecell{84.25 \\ 82.04 \\ 79.66} \\
        \bottomrule
    \end{tabular}
    }
    \vspace{-0.1in}
\end{table}
\begin{figure}
    \centering
    \includegraphics[scale=0.5]{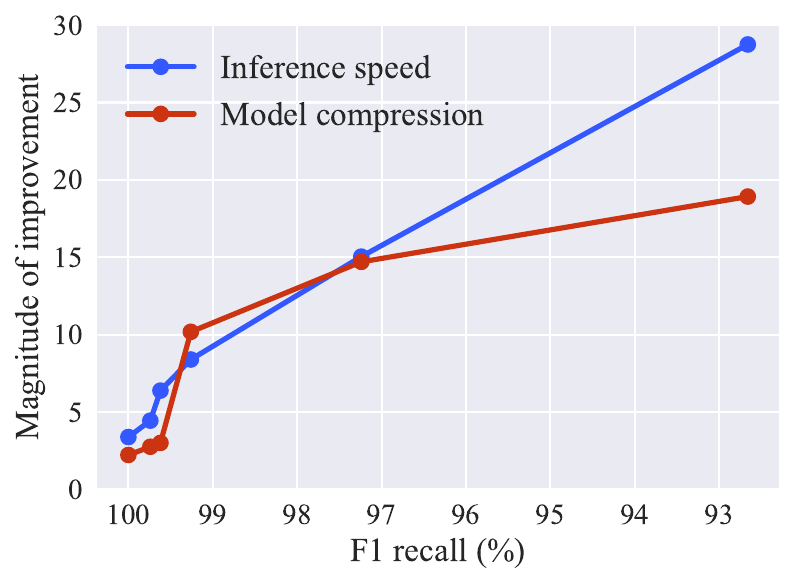}
    \captionof{figure}{F1 recall on the SQuADv1.1 task relative to improvements in CPU-inference speed and model size.}
    \label{fig:inference}
    \vspace{-0.2in}
\end{figure}

\noindent\textbf{Practical trade-offs.} We now benchmark these models in end-to-end fashion, both in terms of model size and inference speed. For model size, we report size of the checkpoint in MB after standard gzip compression. For inference speed, we report number of items per second (throughput) on the well-established SQuAD v1.1 CPU-inference benchmark with a sequence length of 128 and a batch size of 32. Figure~\ref{fig:inference} depicts relative accuracy versus magnitude of improvement in speed and model size. As baseline for full recovery, we follow the community-standard e.g.~\cite{Sanh2020MovementPA}, and adopt the dense \bert model with 88.54 F1 score. The baseline for inference speed is dense \bert inference with DeepSparse, which matches the industry-standard ONNX Runtime inference engine. Results suggest a roughly-linear trade-off between compression and accuracy loss, with a compression jump around 1\% accuracy drop, due to quantization being applied. Specifically, we observe 8.4x higher inference speedup at < 1\% accuracy drop, 10x speedup at < 2\% drop, 15x speedup at < 3\% drop, and 29x speedup at < 7.5\% accuracy drop. This shows how compound compression can optimize LLMs to various latencies. See Appendix Table \ref{tab:deepsparseresults} for full results.

\subsection{Pruning for GPU speedups (N:M sparsity)}
Even though our previous results targeted CPUs for deployment, we now show that our pruning approach can also be relevant to GPUs. We apply the semi-structured variant of oBERT to impose the 2-out-of-4 sparsity pattern, which is supported on NVIDIA Ampere GPUs~\citep{mishra2021accelerating}. More specifically, we prune in \textit{one-shot}, and compare against the magnitude pruning baseline in Table~\ref{tab:gpu24}. All other methods require full fine-tuning, and thus don't support the one-shot setup. oBERT significantly outperforms magnitude pruning, and with only 1-epoch of fine-tuning it is able to fully recover dense accuracy with (F1, EM) = (88.58, 81.16). With this sparsity pattern, the pruned model achieves 1.85x speedup on Ampere devices.

\begin{table}[h!]
    \caption{One-shot 2:4 pruning of the fine-tuned \bert model.}
    \label{tab:gpu24}
    \centering
    {\small
    \begin{tabular}{cc|cc}
    \toprule 
    Task & \makecell{\bert} & \makecell{Magnitude} & oBERT (ours) \\
    \midrule
    \makecell{SQuAD \\ F1 / EM} & 88.54 / 81.41 & 49.97 / 35.24 & \textbf{83.17 / 74.18} \\ 
    \bottomrule
    \end{tabular}
    }
\end{table}

\section{Discussion}

\noindent\textbf{Comparison with concurrent work.} Concurrent work introduced PLATON~\cite{zhang2022platon}, which addresses unstructured pruning of BERT models via estimates of confidence bounds. It does not make use of KD, so for a fair comparison we rerun our experiments without KD as well. Contrary to PLATON, which reports best results after an extensive hyper-parameter search for each task independently, we apply our sparse-transfer setup with the upstream pruned model and only sweep for the number of epochs $\in [1, 8]$. We employ early stopping to prevent overfitting on smaller GLUE tasks. As can be seen from Table~\ref{tab:vs_platon}, oBERT outperforms  PLATON across all tasks.

\begin{table}[h!]
    \caption{Compressed \bert models to 90\% sparsity on GLUE tasks without knowledge distillation.}
    \label{tab:vs_platon}
    \centering
    {\small
    \begin{tabular}{cc|cccccc}
    \toprule 
    Task & \makecell{\bert} &  PLATON & \makecell{oBERT\\(ours)} \\
    \midrule
    \makecell{MNLI\\m / mm} & 84.6 / 83.4 & 82.0 / 82.2 & \textbf{82.2 / 82.5} \\
    \midrule
    \makecell{QQP\\Acc / F1} & 91.5 / 88.5 & 90.2 / 86.8 & \textbf{90.4 / 87.1} \\ 
    \midrule
    \makecell{QNLI\\Acc} & 91.3 & 88.9 & \textbf{89.3} \\ 
    \midrule
    \makecell{MRPC\\Acc / F1} & 86.4 / 90.3 & 84.3 / 88.8 & \textbf{85.6 / 89.3} \\ 
    \midrule
    \makecell{SST-2\\Acc} & 92.7 & 90.5 & \textbf{92.0} \\
    \midrule
    \makecell{CoLA\\Mcc} & 58.3 & 44.3 & \textbf{48.47} \\
    \midrule
    \makecell{STS-B\\Pear / Spear} & 90.2 / 89.7 & 87.4 / 87.1 & \textbf{88.0 / 87.6} \\
    \bottomrule
    \end{tabular}
    }
\end{table}

\noindent\textbf{Broader comparison.} 
We now contrast our compound-compressed \bert models relative to alternative compression techniques. We compare against DistilBERT~\cite{Sanh2019DistilBERTAD}, TinyBERT~\cite{Jiao2020TinyBERTDB}, and Block Pruning For Faster Transformers (Hybrid Filled MvP)~\cite{lagunas21block}. 
DistilBERT leverages KD during pre-training and fine-tuning to obtain a 6-layer model fine-tuned for a specific downstream task. TinyBERT makes use of a specialized Transformer-KD scheme to distill knowledge and intermediate representations at both stages, pre-training and fine-tuning on a specific task. In contrast, we use a simpler approach and employ KD from teacher's outputs only. Hybrid Filled MvP~\cite{lagunas21block} employs semi-structured pruning and weight reintroduction. The comparison is given in Table~\ref{tab:other-methods}, where we report the number of unpruned encoder weights as size, compression ratio and inference speedup relative to the dense \bert in the same inference environment, and F1 score on the dev-set of the SQuAD v1.1 dataset. The results suggest that our compressed models improve upon the current state-of-the-art techniques, setting new very competitive baselines with respect to all metrics: accuracy, model size, and inference speed. 

\begin{table}[h!]
\setlength{\tabcolsep}{4pt}
    \caption{Compressed \bert models on the SQuADv1.1 task. ($\textrm{oBERT}_{6,80}$ stands for the 6-layer model pruned to 80\% sparsity.)}
    \label{tab:other-methods}
    \centering
    {\small 
    \begin{tabular}{lccccc}
    \toprule 
    Model & Size & Compr. & Speedup & F1 & Dev. \\
    \midrule
    \bert & 85.0M & \phantom{1}1.00x & \phantom{1}1.00x & 88.54 &\\
    \midrule
    \multicolumn{5}{l}{\textit{< 6-layers}} \\
    $\textrm{TinyBERT}_4$ & \phantom{8}4.5M & 18.88x & \phantom{1}9.40x & 82.10 & GPU\\
    $\textrm{oBERT}_{3,90}$ & \phantom{8}\textbf{2.1M} & \textbf{40.00x} & \textbf{14.80x} & \textbf{82.50} & CPU\\
    \midrule
    \multicolumn{5}{l}{\textit{6-layers}} \\
    DistilBERT & 42.5M & \phantom{1}2.00x & 2.00x & 86.90 & GPU\\
    $\textrm{TinyBERT}_6$ & 42.5M & \phantom{1}2.00x & 2.00x & 87.50 & GPU\\
    $\textrm{oBERT}_{6,80}$ & \phantom{1}\textbf{8.5M} & \textbf{10.00x} & \textbf{6.38x} & \textbf{88.20} & CPU\\ 
    \midrule
    \multicolumn{5}{l}{\textit{12-layers}} \\
    Hybrid F. MvP & 30.7M & \phantom{1}2.76x & 1.84x & 88.70 & GPU\\
    $\textrm{oBERT}_{12,80}$ & \textbf{17.0M} & \phantom{1}\textbf{5.00x} & \textbf{3.38x} & \textbf{89.04} & CPU\\
    \bottomrule
    \end{tabular}
    }
\end{table}

\noindent\textbf{\bertL results.} Most of our results presented in Section~\ref{sec:experiments} targeted the widely-adopted \bert model. This gave us an opportunity for a fair comparison against many different methods. To verify that our approach does not pertain only to the \bert model, in Table~\ref{tab:bertL} we present downstream pruning results on the three times larger \bertL model and the SQuADv1.1 task. As can be seen from the Table, even the model pruned with oBERT at double the sparsity (95\%) outperforms Prune OFA (90\%).

\begin{table}[h!]
    \caption{Compressed \bertL models on the SQuADv1.1 task.}
    \label{tab:bertL}
    \centering
    {\small
    \begin{tabular}{cc|cc}
    \toprule 
    \makecell{\bertL\\F1 / EM} & Sparsity & \makecell{Prune\\OFA} & \makecell{oBERT \\(ours)} \\
    \midrule
    91.22 / 84.45 & 90\% & 90.20 / 83.35 & \textbf{91.07 / 84.61}\\
    \midrule
    91.22 / 84.45 & 95\% & NA & 90.29 / 83.58\\
    \bottomrule
    \end{tabular}
    }
\end{table}

\noindent\textbf{MLPerf Inference Benchmark.} Motivated by our state-of-the-art results across-the-board, we apply our full compound compression pipeline to compress \bertL and MobileBERT~\cite{sun2020mobilebert} models in the context of the industrial MLPerf Inference Benchmark\footnote{https://mlcommons.org/en/}. In brief, we were able to achieve order-of-magnitude improvements in terms of model size and inference speedups, while maintaining >99\% of the dense \bertL accuracy. For details please see Appendix~\ref{app:mlperf}, as well as our open-source submission.

\section{Broader Impact}
Our work is part of the general trend of producing inference efficient models which approximate performance of their larger bases. By and large, this work should help increase model efficiency, thereby reducing computational and ultimately monetary cost of executing such models. Moreover, it could allow models to be used by those who do not have access to expensive specialized computing clusters: for instance, our main speedup results are aimed at widely-available CPUs. 

\section{Limitations}

As any academic study, our work is not without its limitations. 
We split their discussion into limitations that are  \emph{inherent to our method}, and limitations \emph{of our present study}; the latter can be overcome by extensions of our work.
In the first category, we begin by highlighting the fact that our second-order method relies on approximations, which are inherent in order to scale such methods to BERT scale. 
Prior studies, e.g.~\cite{Singh2020WoodFisherES} have performed careful examinations of the validity of these approximations in the context of CNN models. 
The strength of our empirical results can be seen as indirect evidence that these approximations apply to BERT models as well. 
A second, technical, limitation is the fact that our method requires non-trivial additional storage cost; while we have shown that our experiments can be executed on a single commodity GPU (NVIDIA RTX 3090), this limits the range of devices on which the technique may be applied. However, we provide an efficient and easy way to scale our approach with more GPUs, which is automatically utilized in a multi-GPU environment.  

Another limitation which we aim to remove in future work is the focus on relatively fine-grained sparsity types, such as unstructured and semi-structured pruning.

\bibliographystyle{acl_natbib}
\bibliography{custom}

\begin{thebibliography}{52}
\expandafter\ifx\csname natexlab\endcsname\relax\def\natexlab#1{#1}\fi

\bibitem[{Chen et~al.(2020)Chen, Frankle, Chang, Liu, Zhang, Wang, and
  Carbin}]{chen2020lottery}
Tianlong Chen, Jonathan Frankle, Shiyu Chang, Sijia Liu, Yang Zhang, Zhangyang
  Wang, and Michael Carbin. 2020.
\newblock The lottery ticket hypothesis for pre-trained bert networks.
\newblock \emph{Advances in neural information processing systems},
  33:15834--15846.

\bibitem[{Courbariaux et~al.(2016)Courbariaux, Hubara, Soudry, El-Yaniv, and
  Bengio}]{Courbariaux2016BinarizedNN}
Matthieu Courbariaux, Itay Hubara, Daniel Soudry, Ran El-Yaniv, and Yoshua
  Bengio. 2016.
\newblock Binarized neural networks: Training deep neural networks with weights
  and activations constrained to+ 1 or-1.
\newblock \emph{arXiv preprint arXiv:1602.02830}.

\bibitem[{Devlin et~al.(2019)Devlin, Chang, Lee, and
  Toutanova}]{Devlin2019BERTPO}
Jacob Devlin, Ming-Wei Chang, Kenton Lee, and Kristina Toutanova. 2019.
\newblock {BERT}: Pre-training of deep bidirectional transformers for language
  understanding.
\newblock In \emph{Proceedings of the 2019 Conference of the North {A}merican
  Chapter of the Association for Computational Linguistics: Human Language
  Technologies, Volume 1 (Long and Short Papers)}, pages 4171--4186,
  Minneapolis, Minnesota. Association for Computational Linguistics.

\bibitem[{Fan et~al.(2019)Fan, Grave, and Joulin}]{fan2019reducing}
Angela Fan, Edouard Grave, and Armand Joulin. 2019.
\newblock Reducing transformer depth on demand with structured dropout.
\newblock In \emph{International Conference on Learning Representations}.

\bibitem[{Foundation()}]{wikidump}
Wikimedia Foundation.
\newblock \href {https://dumps.wikimedia.org} {Wikimedia downloads}.

\bibitem[{Frankle and Carbin(2018)}]{Frankle2019TheLT}
Jonathan Frankle and Michael Carbin. 2018.
\newblock The lottery ticket hypothesis: Finding sparse, trainable neural
  networks.
\newblock In \emph{International Conference on Learning Representations}.

\bibitem[{Frantar et~al.(2021)Frantar, Kurtic, and
  Alistarh}]{Frantar2021EfficientMA}
Elias Frantar, Eldar Kurtic, and Dan Alistarh. 2021.
\newblock M-fac: Efficient matrix-free approximations of second-order
  information.
\newblock \emph{Advances in Neural Information Processing Systems}, 34.

\bibitem[{Gale et~al.(2019)Gale, Elsen, and Hooker}]{Gale2019TheSO}
Trevor Gale, Erich Elsen, and Sara Hooker. 2019.
\newblock The state of sparsity in deep neural networks.
\newblock \emph{arXiv preprint arXiv:1902.09574}.

\bibitem[{Han et~al.(2015)Han, Mao, and Dally}]{Han2015ADN}
Song Han, Huizi Mao, and William~J Dally. 2015.
\newblock A deep neural network compression pipeline: Pruning, quantization,
  huffman encoding.
\newblock \emph{arXiv preprint arXiv:1510.00149}, 10.

\bibitem[{Hassibi and Stork(1992)}]{hassibi1993second}
Babak Hassibi and David Stork. 1992.
\newblock Second order derivatives for network pruning: Optimal brain surgeon.
\newblock \emph{Advances in neural information processing systems}, 5.

\bibitem[{Hinton et~al.(2015)Hinton, Vinyals, Dean
  et~al.}]{Hinton2015DistillingTK}
Geoffrey Hinton, Oriol Vinyals, Jeff Dean, et~al. 2015.
\newblock Distilling the knowledge in a neural network.
\newblock \emph{arXiv preprint arXiv:1503.02531}, 2(7).

\bibitem[{Jacob et~al.(2018)Jacob, Kligys, Chen, Zhu, Tang, Howard, Adam, and
  Kalenichenko}]{jacob2018quantization}
Benoit Jacob, Skirmantas Kligys, Bo~Chen, Menglong Zhu, Matthew Tang, Andrew
  Howard, Hartwig Adam, and Dmitry Kalenichenko. 2018.
\newblock Quantization and training of neural networks for efficient
  integer-arithmetic-only inference.
\newblock In \emph{Proceedings of the IEEE conference on computer vision and
  pattern recognition}, pages 2704--2713.

\bibitem[{Jiao et~al.(2020)Jiao, Yin, Shang, Jiang, Chen, Li, Wang, and
  Liu}]{Jiao2020TinyBERTDB}
Xiaoqi Jiao, Yichun Yin, Lifeng Shang, Xin Jiang, Xiao Chen, Linlin Li, Fang
  Wang, and Qun Liu. 2020.
\newblock Tinybert: Distilling bert for natural language understanding.
\newblock In \emph{Findings of the Association for Computational Linguistics:
  EMNLP 2020}, pages 4163--4174.

\bibitem[{Kim et~al.(2022)Kim, Shen, Thorsley, Gholami, Kwon, Hassoun, and
  Keutzer}]{kim2021learned}
Sehoon Kim, Sheng Shen, David Thorsley, Amir Gholami, Woosuk Kwon, Joseph
  Hassoun, and Kurt Keutzer. 2022.
\newblock Learned token pruning for transformers.
\newblock In \emph{Proceedings of the 28th ACM SIGKDD Conference on Knowledge
  Discovery and Data Mining}, KDD '22, page 784–794. Association for
  Computing Machinery.

\bibitem[{Kovaleva et~al.(2021)Kovaleva, Kulshreshtha, Rogers, and
  Rumshisky}]{DBLP:journals/corr/abs-2105-06990}
Olga Kovaleva, Saurabh Kulshreshtha, Anna Rogers, and Anna Rumshisky. 2021.
\newblock {BERT} busters: Outlier layernorm dimensions that disrupt {BERT}.
\newblock \emph{CoRR}, abs/2105.06990.

\bibitem[{Kurtz et~al.(2020)Kurtz, Kopinsky, Gelashvili, Matveev, Carr, Goin,
  Leiserson, Moore, Nell, Shavit, and Alistarh}]{pmlr-v119-kurtz20a}
Mark Kurtz, Justin Kopinsky, Rati Gelashvili, Alexander Matveev, John Carr,
  Michael Goin, William Leiserson, Sage Moore, Bill Nell, Nir Shavit, and Dan
  Alistarh. 2020.
\newblock Inducing and exploiting activation sparsity for fast inference on
  deep neural networks.
\newblock In \emph{Proceedings of the 37th International Conference on Machine
  Learning}, volume 119 of \emph{Proceedings of Machine Learning Research},
  pages 5533--5543, Virtual. PMLR.

\bibitem[{Lagunas et~al.(2021)Lagunas, Charlaix, Sanh, and
  Rush}]{lagunas21block}
Fran{\c{c}}ois Lagunas, Ella Charlaix, Victor Sanh, and Alexander Rush. 2021.
\newblock Block pruning for faster transformers.
\newblock In \emph{Proceedings of the 2021 Conference on Empirical Methods in
  Natural Language Processing}, pages 10619--10629, Online and Punta Cana,
  Dominican Republic. Association for Computational Linguistics.

\bibitem[{LeCun et~al.(1989)LeCun, Denker, and Solla}]{LeCun1989OptimalBD}
Yann LeCun, John Denker, and Sara Solla. 1989.
\newblock Optimal brain damage.
\newblock \emph{Advances in neural information processing systems}, 2.

\bibitem[{Lhoest et~al.(2021)Lhoest, Villanova~del Moral, Jernite, Thakur, von
  Platen, Patil, Chaumond, Drame, Plu, Tunstall, Davison, {\v{S}}a{\v{s}}ko,
  Chhablani, Malik, Brandeis, Le~Scao, Sanh, Xu, Patry, McMillan-Major, Schmid,
  Gugger, Delangue, Matussi{\`e}re, Debut, Bekman, Cistac, Goehringer, Mustar,
  Lagunas, Rush, and Wolf}]{hf-datasets}
Quentin Lhoest, Albert Villanova~del Moral, Yacine Jernite, Abhishek Thakur,
  Patrick von Platen, Suraj Patil, Julien Chaumond, Mariama Drame, Julien Plu,
  Lewis Tunstall, Joe Davison, Mario {\v{S}}a{\v{s}}ko, Gunjan Chhablani,
  Bhavitvya Malik, Simon Brandeis, Teven Le~Scao, Victor Sanh, Canwen Xu,
  Nicolas Patry, Angelina McMillan-Major, Philipp Schmid, Sylvain Gugger,
  Cl{\'e}ment Delangue, Th{\'e}o Matussi{\`e}re, Lysandre Debut, Stas Bekman,
  Pierric Cistac, Thibault Goehringer, Victor Mustar, Fran{\c{c}}ois Lagunas,
  Alexander Rush, and Thomas Wolf. 2021.
\newblock Datasets: A community library for natural language processing.
\newblock In \emph{Proceedings of the 2021 Conference on Empirical Methods in
  Natural Language Processing: System Demonstrations}, pages 175--184.
  Association for Computational Linguistics.

\bibitem[{Liu et~al.(2019)Liu, Ott, Goyal, Du, Joshi, Chen, Levy, Lewis,
  Zettlemoyer, and Stoyanov}]{Liu2019RoBERTaAR}
Yinhan Liu, Myle Ott, Naman Goyal, Jingfei Du, Mandar Joshi, Danqi Chen, Omer
  Levy, Mike Lewis, Luke Zettlemoyer, and Veselin Stoyanov. 2019.
\newblock Roberta: A robustly optimized bert pretraining approach.
\newblock \emph{arXiv preprint arXiv:1907.11692}.

\bibitem[{Michel et~al.(2019)Michel, Levy, and Neubig}]{Michel2019AreSH}
Paul Michel, Omer Levy, and Graham Neubig. 2019.
\newblock Are sixteen heads really better than one?
\newblock \emph{Advances in neural information processing systems}, 32.

\bibitem[{Mishra et~al.(2021)Mishra, Latorre, Pool, Stosic, Stosic, Venkatesh,
  Yu, and Micikevicius}]{mishra2021accelerating}
Asit Mishra, Jorge~Albericio Latorre, Jeff Pool, Darko Stosic, Dusan Stosic,
  Ganesh Venkatesh, Chong Yu, and Paulius Micikevicius. 2021.
\newblock Accelerating sparse deep neural networks.
\newblock \emph{arXiv preprint arXiv:2104.08378}.

\bibitem[{NeuralMagic(2021)}]{deepsparse}
NeuralMagic. 2021.
\newblock \href
  {http://arxiv.org/abs/https://github.com/neuralmagic/deepsparse} {Deep
  sparse: A fast cpu inference engine}.

\bibitem[{Radford et~al.(2019)Radford, Wu, Child, Luan, Amodei, Sutskever
  et~al.}]{Radford2019LanguageMA}
Alec Radford, Jeffrey Wu, Rewon Child, David Luan, Dario Amodei, Ilya
  Sutskever, et~al. 2019.
\newblock Language models are unsupervised multitask learners.
\newblock \emph{OpenAI blog}, 1(8):9.

\bibitem[{Rajpurkar et~al.(2016)Rajpurkar, Zhang, Lopyrev, and
  Liang}]{Rajpurkar2016SQuAD1Q}
Pranav Rajpurkar, Jian Zhang, Konstantin Lopyrev, and Percy Liang. 2016.
\newblock Squad: 100,000+ questions for machine comprehension of text.
\newblock In \emph{EMNLP}.

\bibitem[{Sajjad et~al.(2020)Sajjad, Dalvi, Durrani, and
  Nakov}]{DBLP:journals/corr/abs-2004-03844}
Hassan Sajjad, Fahim Dalvi, Nadir Durrani, and Preslav Nakov. 2020.
\newblock Poor man's {BERT:} smaller and faster transformer models.
\newblock \emph{CoRR}, abs/2004.03844.

\bibitem[{Sanh et~al.(2019)Sanh, Debut, Chaumond, and
  Wolf}]{Sanh2019DistilBERTAD}
Victor Sanh, Lysandre Debut, Julien Chaumond, and Thomas Wolf. 2019.
\newblock Distilbert, a distilled version of bert: smaller, faster, cheaper and
  lighter.
\newblock \emph{arXiv preprint arXiv:1910.01108}.

\bibitem[{Sanh et~al.(2020)Sanh, Wolf, and Rush}]{Sanh2020MovementPA}
Victor Sanh, Thomas Wolf, and Alexander Rush. 2020.
\newblock Movement pruning: Adaptive sparsity by fine-tuning.
\newblock \emph{Advances in Neural Information Processing Systems},
  33:20378--20389.

\bibitem[{Shankar et~al.(2017)Shankar, Nikhil, and Kornel}]{shankar2017first}
Iyer Shankar, Dandekar Nikhil, and Csernai Kornel. 2017.
\newblock First quora dataset release: Question pairs.

\bibitem[{Shankar(2017)}]{Shankar2017IdentifyingQQ}
S.~Shankar. 2017.
\newblock Identifying quora question pairs having the same intent.

\bibitem[{Shen et~al.(2020)Shen, Dong, Ye, Ma, Yao, Gholami, Mahoney, and
  Keutzer}]{shen2020q}
Sheng Shen, Zhen Dong, Jiayu Ye, Linjian Ma, Zhewei Yao, Amir Gholami,
  Michael~W Mahoney, and Kurt Keutzer. 2020.
\newblock Q-bert: Hessian based ultra low precision quantization of bert.
\newblock In \emph{Proceedings of the AAAI Conference on Artificial
  Intelligence}, volume~34, pages 8815--8821.

\bibitem[{Singh and Alistarh(2020)}]{Singh2020WoodFisherES}
Sidak~Pal Singh and Dan Alistarh. 2020.
\newblock Woodfisher: Efficient second-order approximation for neural network
  compression.
\newblock \emph{Advances in Neural Information Processing Systems}, 33.

\bibitem[{Smith et~al.(2022)Smith, Patwary, Norick, LeGresley, Rajbhandari,
  Casper, Liu, Prabhumoye, Zerveas, Korthikanti et~al.}]{MTNLG}
Shaden Smith, Mostofa Patwary, Brandon Norick, Patrick LeGresley, Samyam
  Rajbhandari, Jared Casper, Zhun Liu, Shrimai Prabhumoye, George Zerveas,
  Vijay Korthikanti, et~al. 2022.
\newblock Using deepspeed and megatron to train megatron-turing nlg 530b, a
  large-scale generative language model.
\newblock \emph{arXiv preprint arXiv:2201.11990}.

\bibitem[{Sridhar and Sarah(2020)}]{Sridhar2020UndividedAA}
Sharath~Nittur Sridhar and Anthony Sarah. 2020.
\newblock Undivided attention: Are intermediate layers necessary for bert?
\newblock \emph{arXiv preprint arXiv:2012.11881}.

\bibitem[{Sun et~al.(2020{\natexlab{a}})Sun, Yu, Song, Liu, Yang, and
  Zhou}]{Sun2020MobileBERTAC}
Zhiqing Sun, Hongkun Yu, Xiaodan Song, Renjie Liu, Yiming Yang, and Denny Zhou.
  2020{\natexlab{a}}.
\newblock Mobilebert: a compact task-agnostic bert for resource-limited
  devices.
\newblock In \emph{ACL}.

\bibitem[{Sun et~al.(2020{\natexlab{b}})Sun, Yu, Song, Liu, Yang, and
  Zhou}]{sun2020mobilebert}
Zhiqing Sun, Hongkun Yu, Xiaodan Song, Renjie Liu, Yiming Yang, and Denny Zhou.
  2020{\natexlab{b}}.
\newblock Mobilebert: a compact task-agnostic bert for resource-limited
  devices.
\newblock In \emph{Proceedings of the 58th Annual Meeting of the Association
  for Computational Linguistics}, pages 2158--2170.

\bibitem[{Turc et~al.(2019)Turc, Chang, Lee, and
  Toutanova}]{Turc2019WellReadSL}
Iulia Turc, Ming-Wei Chang, Kenton Lee, and Kristina Toutanova. 2019.
\newblock Well-read students learn better: The impact of student initialization
  on knowledge distillation.
\newblock \emph{ArXiv}, abs/1908.08962.

\bibitem[{Vaswani et~al.(2017)Vaswani, Shazeer, Parmar, Uszkoreit, Jones,
  Gomez, Kaiser, and Polosukhin}]{Vaswani2017AttentionIA}
Ashish Vaswani, Noam Shazeer, Niki Parmar, Jakob Uszkoreit, Llion Jones,
  Aidan~N Gomez, {\L}ukasz Kaiser, and Illia Polosukhin. 2017.
\newblock Attention is all you need.
\newblock \emph{Advances in neural information processing systems}, 30.

\bibitem[{Voita et~al.(2019)Voita, Talbot, Moiseev, Sennrich, and
  Titov}]{Voita2019AnalyzingMS}
Elena Voita, David Talbot, F.~Moiseev, Rico Sennrich, and Ivan Titov. 2019.
\newblock Analyzing multi-head self-attention: Specialized heads do the heavy
  lifting, the rest can be pruned.
\newblock In \emph{ACL}.

\bibitem[{Wang et~al.(2020)Wang, Wei, Dong, Bao, Yang, and
  Zhou}]{Wang2020MiniLMDS}
Wenhui Wang, Furu Wei, Li~Dong, Hangbo Bao, Nan Yang, and Ming Zhou. 2020.
\newblock Minilm: Deep self-attention distillation for task-agnostic
  compression of pre-trained transformers.
\newblock \emph{Advances in Neural Information Processing Systems},
  33:5776--5788.

\bibitem[{Weijie et~al.(2020)Weijie, Peng, Zhe, Zhiruo, Haotang, and
  Qi}]{liu2020fastbert}
Liu Weijie, Zhou Peng, Zhao Zhe, Wang Zhiruo, Deng Haotang, and Ju~Qi. 2020.
\newblock Fastbert: a self-distilling bert with adaptive inference time.
\newblock In \emph{Proceedings of ACL 2020}.

\bibitem[{Williams et~al.(2018)Williams, Nangia, and Bowman}]{N18-1101}
Adina Williams, Nikita Nangia, and Samuel Bowman. 2018.
\newblock A broad-coverage challenge corpus for sentence understanding through
  inference.
\newblock In \emph{Proceedings of the 2018 Conference of the North American
  Chapter of the Association for Computational Linguistics: Human Language
  Technologies, Volume 1 (Long Papers)}, pages 1112--1122. Association for
  Computational Linguistics.

\bibitem[{Wolf et~al.(2020)Wolf, Debut, Sanh, Chaumond, Delangue, Moi, Cistac,
  Rault, Louf, Funtowicz, Davison, Shleifer, von Platen, Ma, Jernite, Plu, Xu,
  Scao, Gugger, Drame, Lhoest, and Rush}]{wolf-etal-2020-transformers}
Thomas Wolf, Lysandre Debut, Victor Sanh, Julien Chaumond, Clement Delangue,
  Anthony Moi, Pierric Cistac, Tim Rault, Rémi Louf, Morgan Funtowicz, Joe
  Davison, Sam Shleifer, Patrick von Platen, Clara Ma, Yacine Jernite, Julien
  Plu, Canwen Xu, Teven~Le Scao, Sylvain Gugger, Mariama Drame, Quentin Lhoest,
  and Alexander~M. Rush. 2020.
\newblock Transformers: State-of-the-art natural language processing.
\newblock In \emph{Proceedings of the 2020 Conference on Empirical Methods in
  Natural Language Processing: System Demonstrations}, pages 38--45, Online.
  Association for Computational Linguistics.

\bibitem[{Xin et~al.(2020)Xin, Tang, Lee, Yu, and Lin}]{Xin2020DeeBERTDE}
Ji~Xin, Raphael Tang, Jaejun Lee, Yaoliang Yu, and Jimmy~J. Lin. 2020.
\newblock Deebert: Dynamic early exiting for accelerating bert inference.
\newblock In \emph{ACL}.

\bibitem[{Xu et~al.(2021)Xu, Yen, Zhao, and Xiao}]{Xu2021RethinkingNP}
Dongkuan Xu, Ian En-Hsu Yen, Jinxi Zhao, and Zhibin Xiao. 2021.
\newblock Rethinking network pruning – under the pre-train and fine-tune
  paradigm.
\newblock In \emph{NAACL}.

\bibitem[{Yu et~al.(2022)Yu, Yao, Gholami, Dong, Kim, Mahoney, and
  Keutzer}]{yu2022hessian}
Shixing Yu, Zhewei Yao, Amir Gholami, Zhen Dong, Sehoon Kim, Michael~W Mahoney,
  and Kurt Keutzer. 2022.
\newblock Hessian-aware pruning and optimal neural implant.
\newblock In \emph{Proceedings of the IEEE/CVF Winter Conference on
  Applications of Computer Vision}, pages 3880--3891.

\bibitem[{Zafrir et~al.(2019)Zafrir, Boudoukh, Izsak, and
  Wasserblat}]{Zafrir2019Q8BERTQ8}
Ofir Zafrir, Guy Boudoukh, Peter Izsak, and Moshe Wasserblat. 2019.
\newblock Q8bert: Quantized 8bit bert.
\newblock \emph{2019 Fifth Workshop on Energy Efficient Machine Learning and
  Cognitive Computing - NeurIPS Edition (EMC2-NIPS)}, pages 36--39.

\bibitem[{Zafrir et~al.(2021)Zafrir, Larey, Boudoukh, Shen, and
  Wasserblat}]{zafrir2021prune}
Ofir Zafrir, Ariel Larey, Guy Boudoukh, Haihao Shen, and Moshe Wasserblat.
  2021.
\newblock Prune once for all: Sparse pre-trained language models.
\newblock \emph{arXiv preprint arXiv:2111.05754}.

\bibitem[{Zhang et~al.(2022)Zhang, Zuo, Liang, Bukharin, He, Chen, and
  Zhao}]{zhang2022platon}
Qingru Zhang, Simiao Zuo, Chen Liang, Alexander Bukharin, Pengcheng He, Weizhu
  Chen, and Tuo Zhao. 2022.
\newblock Platon: Pruning large transformer models with upper confidence bound
  of weight importance.
\newblock In \emph{International Conference on Machine Learning}, pages
  26809--26823. PMLR.

\bibitem[{Zhang et~al.(2020)Zhang, Hou, Yin, Shang, Chen, Jiang, and
  Liu}]{zhang2020ternarybert}
Wei Zhang, Lu~Hou, Yichun Yin, Lifeng Shang, Xiao Chen, Xin Jiang, and Qun Liu.
  2020.
\newblock Ternarybert: Distillation-aware ultra-low bit bert.
\newblock \emph{arXiv preprint arXiv:2009.12812}.

\bibitem[{Zhu and Gupta(2018)}]{Zhu2018ToPO}
M.~Zhu and Suyog Gupta. 2018.
\newblock To prune, or not to prune: exploring the efficacy of pruning for
  model compression.
\newblock \emph{ArXiv}, abs/1710.01878.

\bibitem[{Zhu et~al.(2015)Zhu, Kiros, Zemel, Salakhutdinov, Urtasun, Torralba,
  and Fidler}]{Zhu2015AligningBA}
Yukun Zhu, Ryan Kiros, Richard~S. Zemel, Ruslan Salakhutdinov, Raquel Urtasun,
  Antonio Torralba, and Sanja Fidler. 2015.
\newblock Aligning books and movies: Towards story-like visual explanations by
  watching movies and reading books.
\newblock \emph{2015 IEEE International Conference on Computer Vision (ICCV)},
  pages 19--27.

\end{thebibliography}


\appendix
\section{Appendix}
\label{sec:appendix}

\subsection{MLPerf Inference benchmark}
\label{app:mlperf}
Following the MLPerf benchmark guidelines on producing compressed and fast models while maintaining >99\% of the \bertL F1 score on the SQuADv1.1 task, we explore two directions. In the first one, dubbed oBERT-Large, we compound compress the \bertL model without any changes to its architecture. Therefore, we apply 4-block downstream pruning to 95\% sparsity followed by the quantization aware training (QAT). In the second direction we focus on recovering the \bertL accuracy by compressing an already compact MobileBERT model, dubbed oBERT-MobileBERT. More specifically, we apply direct layer dropping, leaving only 14 transformer layers out of the original 24, followed by the 4-block pruning to 50\% sparsity and quantization aware training. We present results in Table~\ref{tab:mlperf}, where models were evaluated with the DeepSparse inference engine, using a server with two Intel(R) Xeon(R) Platinum 8380 (IceLake) CPUs with 40 cores each, batch-size 128 and sequence length 384. For more details please see our official submission at \url{https://github.com/neuralmagic/mlperf_inference_results_v2.1/tree/master/open/NeuralMagic}.

\begin{table*}[ht!]
    \caption{MLPerf inference results for oBERT compressed \bertL and MobileBERT models.}
    \label{tab:mlperf}
    \centering
    {\small
    \begin{tabular}{ccccccccc}
    \toprule 
    Model & Precision & \makecell{F1 Score\\(R=X\% recovery)} & File Size & \makecell{Compression\\Ratio} & \makecell{Throughput\\(samples/sec)} & Speedup \\
    \midrule
    \makecell{BERT-Large\\dense baseline} & FP32 & 90.87 (R=100\%) & 1.30 GB & 1x & 15.49 & 1x \\
    \midrule
    oBERT-Large & INT8 & 90.21 (R=99.27\%) & 38.20 MB & \phantom{1}34x & 230.74 & 15x \\
    oBERT-MobileBERT & INT8 & 90.32 (R=99.39\%) & \phantom{1}9.56 MB & 136x & 928.58 & 60x \\
    \bottomrule
    \end{tabular}
    }
\end{table*}

\subsection{Additional comparisons}
\label{app:additional_comparisons}
Here we reflect upon some other methods focused on efficient inference for LLMs, which are orthogonal to weight pruning. For example, Learned Token Pruning~\cite{kim2021learned} tries to adaptively remove unimportant tokens in input sequences and provides 2x higher throughput at < 1\% accuracy drop; at the same accuracy drop, our compressed model is able to achieve 8.4x higher throughput. DeeBERT~\cite{Xin2020DeeBERTDE} and FastBERT~\cite{liu2020fastbert} apply an \emph{early-exit} technique for inference speedup. The latter achieves 2-3x faster inference without performance degradation. However, the method only applies to batch size one. Nevertheless, in terms of direct comparison, our compressed models are able to achieve 4x faster inference on CPUs without accuracy degradation. 
Overall, we emphasize the fact that these methods are complementary to our compression techniques, so it would be interesting to investigate computational gains by combining such methods.

\subsection{Computational costs}
\label{app:computational_costs}
In practice, for the 12-layer \bert model with $d=85M$ encoder weights and block size $B=50$, the $\mathcal{O}(Bd)$ memory requirement translates to approximately 17GB, which can be easily kept on the 24GB RTX 3090 card. While this amount of memory is available on high-performance GPUs, it is also straightforward to split the $N_B \times B \times B$ tensor along the batch-dimension $N_B$ and utilize additional GPUs or even memory swapping with CPU. Our implementation updates the inverse Hessian approximation in negligible time, and can run asynchronously while the next gradient is being fetched. Computing saliency scores and optimal weight updates takes only a few seconds.

\subsection{Optimal BERT Surgeon (oBERT) hyper-parameters}
\label{app:obs-hyperparams}
\noindent\textbf{Hyper-parameters.} The oBERT pruning method has three tunable hyper-parameters: number of gradients ($m$), block size ($B$), and dampening ($\lambda$). These are supposed to be tuned with respect to the model and available computational resources. In all of our runs, across all models and datasets, we use the same set of hyper-parameters which we found to work best for the \bert model on the SQuAD v1.1 dataset. We conjecture that further tuning for smaller models (3 and 6-layer models) could improve their results, but for simplicity and fairness to other methods, we apply the same ones found for the \bert.

\begin{figure}
    \centering
    \includegraphics[scale=0.5]{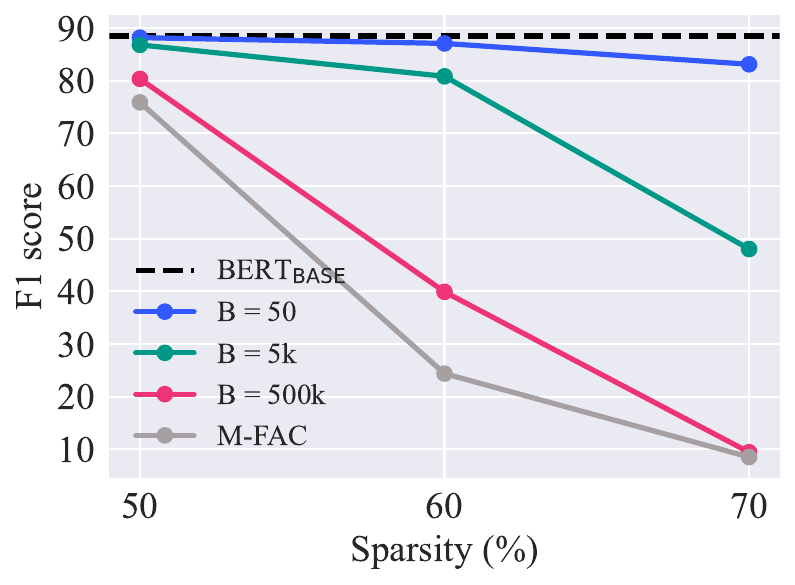}
    \caption{One-shot pruning ablation study with respect to the block size ($B$), with $m = 1024$ and $\lambda = 10^{-7}$, on the \bert model and the question-answering SQuAD v1.1 dataset. M-FAC stands for the full inverse Hessian approximation~\cite{Frantar2021EfficientMA}.}
    \label{fig:ablation_B}
\end{figure}

\begin{figure}
    \centering
    \includegraphics[scale=0.5]{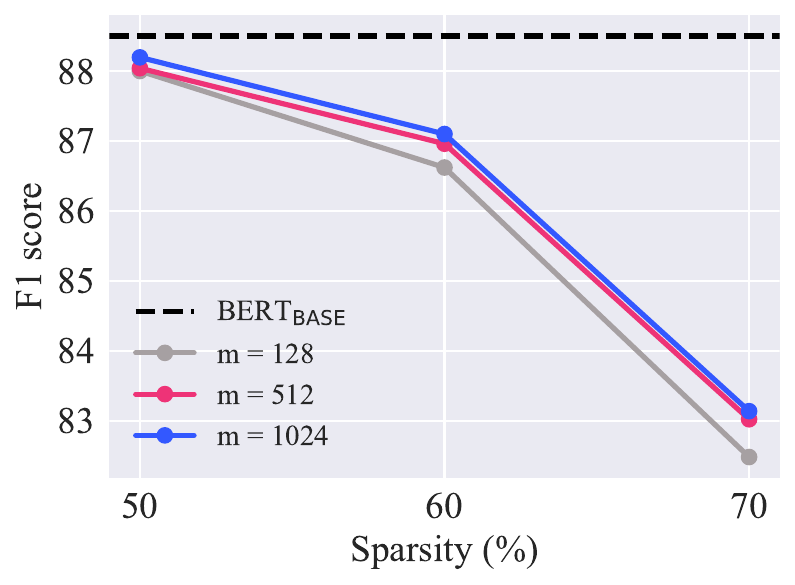}
    \caption{One-shot pruning ablation study with respect to the number of gradients ($m$), with $B = 50$ and $\lambda = 10^{-7}$, on the \bert model and the question-answering SQuAD v1.1 dataset.}
    \label{fig:ablation_m}
\end{figure}

\begin{figure}
    \centering
    \includegraphics[scale=0.5]{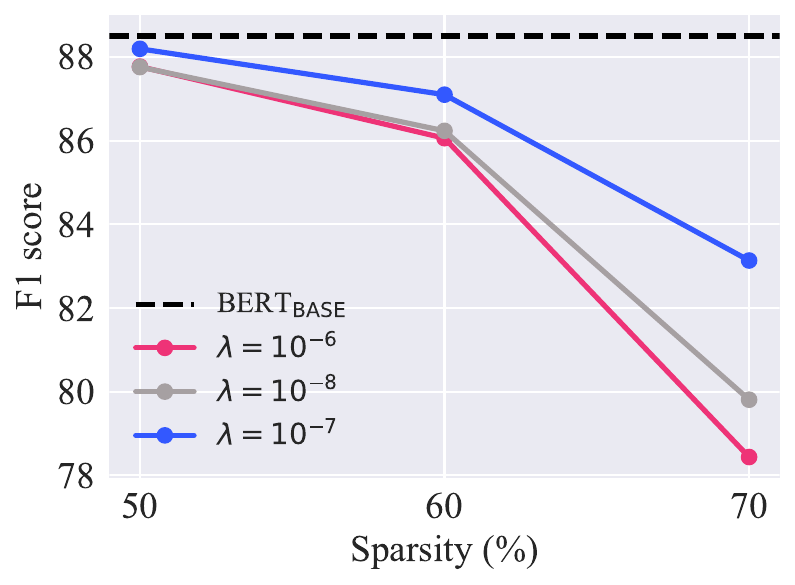}
    \caption{One-shot pruning ablation study with respect to the dampening ($\lambda$), with $m = 1024$ and $B = 50$, on the \bert model and the question-answering SQuAD v1.1 dataset.}
    \label{fig:ablation_damp}
\end{figure}

\noindent\textbf{Ablation studies.} The procedure to find the optimal set of hyper-parameters for a model consists of a grid search over the possible hyper-parameter combinations and one-shot pruning runs to various high sparsity targets to evaluate the quality of the pruning approximation for each combination. We found that $m=1024$, $B=50$, and $\lambda=10^{-7}$ produce state-of-the-art results for a negligible computational overhead with the \bert model. \citet{Frantar2021EfficientMA} shows that larger block sizes require more gradients for better approximation. Given the massive size of the \bert model, we picked this setup as it was the best performing one that could still fit on a single 24GB RTX 3090 GPU card. In Figures \ref{fig:ablation_B}, \ref{fig:ablation_m}, and \ref{fig:ablation_damp} we visualize a fraction of the one-shot pruning ablations with respect to all three hyper-parameters that motivated us to pick these specific values.

\subsection{Downstream pruning}
\label{app:hyperparams-DownstreamPruning}

\noindent\textbf{Teacher preparation.} For all downstream pruning runs we make use of the KD from the fine-tuned \bert teacher outputs. The teacher is fine-tuned on the corresponding downstream task following the default hyper-parameters for SQuAD\footnote{https://github.com/huggingface/transformers/tree/main/\\examples/pytorch/question-answering} and GLUE (QQP and MNLI)\footnote{https://github.com/huggingface/transformers/tree/main/\\examples/pytorch/text-classification}. 

\begin{figure}[htb!]
    \centering
    \includegraphics[scale=0.5]{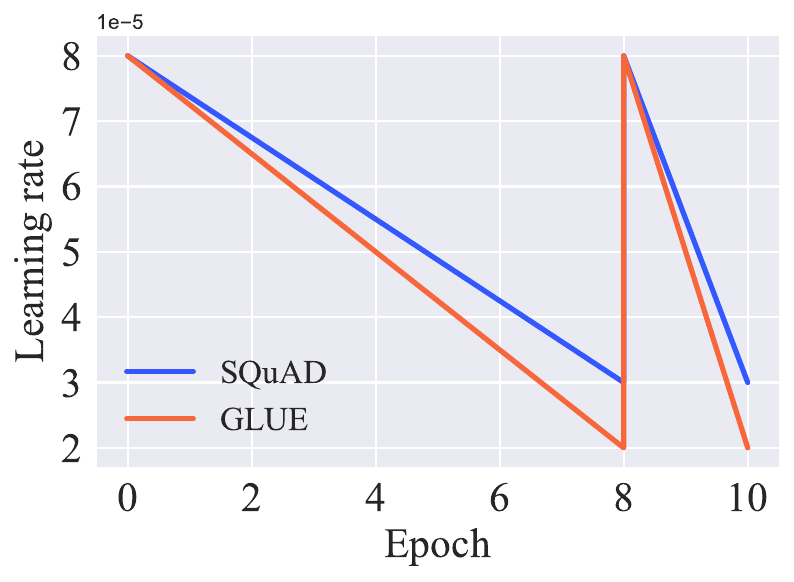}
    \caption{Visualized learning rate schedule for 10-epoch downstream runs.}
    \label{fig:lr_10ep}
\end{figure}

\begin{figure}[htb!]
    \centering
    \includegraphics[scale=0.5]{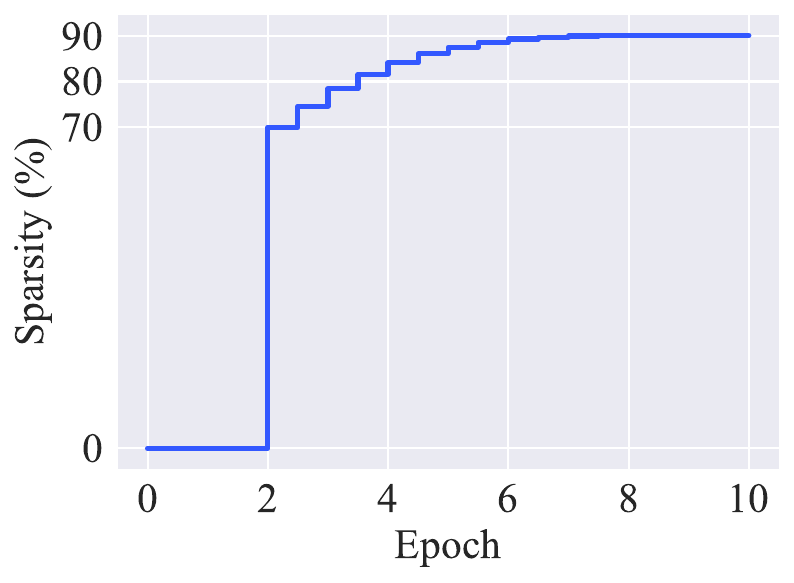}
    \caption{Visualized sparsity schedule for 10-epoch downstream runs with initial sparsity of 70\% and target sparsity of 90\%, following the cubic interpolation~\cite{Zhu2018ToPO}.}
    \label{fig:spars90_ep10}
\end{figure}

\begin{figure}[htb!]
    \centering
    \includegraphics[scale=0.5]{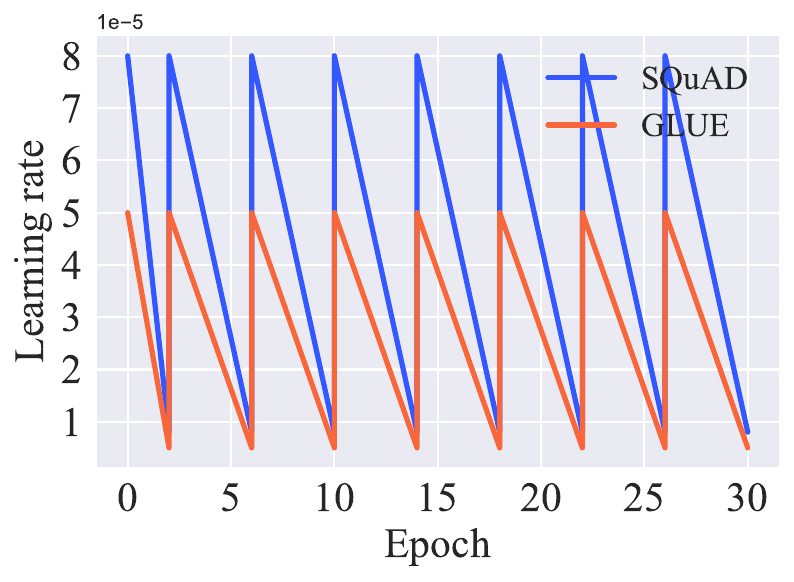}
    \caption{Visualized learning rate schedule for 30-epoch downstream runs.}
    \label{fig:lr_30ep}
\end{figure}

\begin{figure}[htb!]
    \centering
    \includegraphics[scale=0.5]{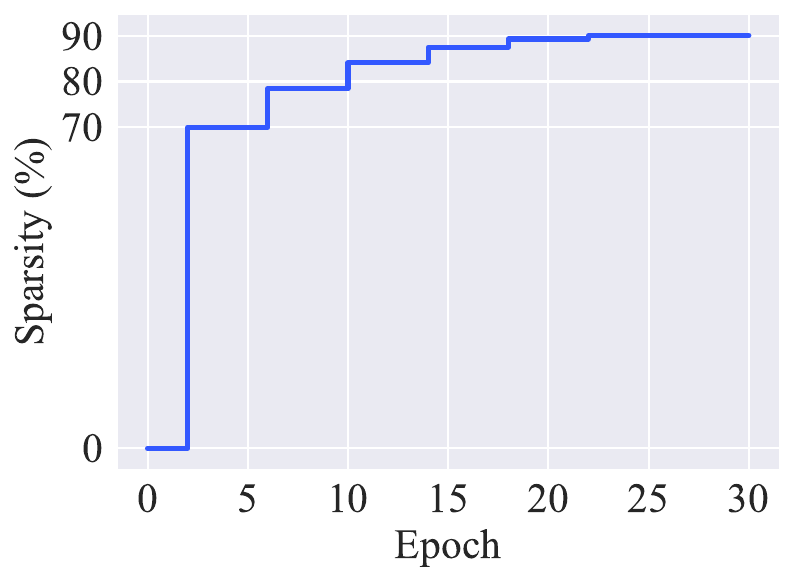}
    \caption{Visualized sparsity schedule for 30-epoch downstream runs with initial sparsity of 70\% and target sparsity of 90\%, following the cubic interpolation~\cite{Zhu2018ToPO}.}
    \label{fig:spars90_ep30}
\end{figure}

\noindent\textbf{Pruning setup.} In Table \ref{tab:hyperparams-DownstreamPruning} we describe in detail all hyper-parameters for downstream pruning results presented in Tables \ref{tab:10_30_gradual} and \ref{tab:downstream-block}. For easier comprehension, we also visualize learning rate schedules in Figures \ref{fig:lr_10ep} and \ref{fig:lr_30ep}, and sparsity schedules in Figures \ref{fig:spars90_ep10} and \ref{fig:spars90_ep30}. 

\noindent\textbf{3-, 6-layer models.} We prepare our 3 and 6 layer models for downstream runs in two stages: layer dropping and retraining phase. We drop layers from our upstream teacher model (more details on it in Appendix \ref{app:hyperparams-UpstreamPruning}). After dropping, we retrain the remaining layers, following insights from \cite{Turc2019WellReadSL}, in the same setup used to prepare the upstream teacher with addition of the KD from it.

\begin{table*}
    \centering
    {\small
    \begin{tabular}{l|cc}
    \toprule
    & 10 Epochs & 30 Epochs \\
    \midrule
    Batch size & \multicolumn{2}{c}{\phantom{xxxxxxxxxxx}\makecell[l]{16 for SQuAD,\\ 32 for GLUE}} \\
    \greyrule
    Learning rate (initial, final) & \makecell[l]{(8e-5, 3e-5) for SQuAD,\\(8e-5, 2e-5) for GLUE} & \makecell[l]{(8e-5, 8e-6) for SQuAD,\\(5e-5, 5e-6) for GLUE} \\
    \greyrule
    Learning rate schedule & \multicolumn{2}{c}{linear decay with rewinds} \\
    \greyrule
    Learning rate rewinds & \makecell{one at epoch=8} & \makecell{periodic every 4 epochs,\\ start at epoch=2} \\
    \greyrule 
    Knowledge Distillation (hardness, temp.) & \multicolumn{2}{c}{(1.0, 2.0)} \\
    \greyrule
    Student model & \multicolumn{2}{c}{\phantom{xxx}\makecell[l]{12-layer: bert-base-uncased\\\phantom{1}6-layer: layer drop + pre-train with KD\\\phantom{1}3-layer: layer drop + pre-train with KD}} \\
    \greyrule
    Teacher model & \multicolumn{2}{c}{\bert} \\
    \greyrule
    Prune start & \multicolumn{2}{c}{epoch=2} \\
    \greyrule
    Prune end & epoch=8 & epoch=26 \\
    \greyrule
    Pruning frequency & 2x per epoch & once every 4 epochs \\
    \greyrule
    Initial sparsity step & \multicolumn{2}{c}{\phantom{xxx}\makecell[l]{12-layer: 70\%\\\phantom{1}6-layer: 30\%\\\phantom{1}3-layer: 30\%}} \\
    \greyrule
    Sparsity distribution & \multicolumn{2}{c}{global over all layers} \\
    \greyrule
    oBERT parameters & \multicolumn{2}{c}{\phantom{xxxxxxxxx}\makecell[l]{Number of gradients $m = 1024$\\ Block size $B=50$\\ Dampening $\lambda=10^{-7}$}}\\
    \bottomrule
    \end{tabular}
    }
    \caption{Downstream pruning hyper-parameters used to obtain results presented in Tables \ref{tab:10_30_gradual} and \ref{tab:downstream-block}.}
    \label{tab:hyperparams-DownstreamPruning}
\end{table*}

\subsection{Upstream pruning}
\label{app:hyperparams-UpstreamPruning}

\noindent\textbf{Teacher preparation.} We prepare a teacher for upstream pruning by following some insights from \cite{Liu2019RoBERTaAR}. More concretely we start with the \textit{bert-base-uncased}\footnote{https://huggingface.co/bert-base-uncased} model, adopt pre-training on two datasets (BookCorpus\footnote{https://huggingface.co/datasets/bookcorpus} \& English Wikipedia\footnote{https://huggingface.co/datasets/wikipedia}) with focus on the masked language modeling task (MLM) for 10-epochs with batch size 256 and learning rate linearly decaying to zero from the initial value of 1e-4. 

\noindent\textbf{Pruning setup.} In Table \ref{tab:hyperparams-UpstreamPruning} we describe in detail our upstream pruning recipe. As can be noticed, our upstream pruning recipe is just a downscaled version of our 30-epoch downstream-pruning recipe to 3-epochs.

\begin{table}
      \centering
         {\small 
            \begin{tabular}{l|c}
            \toprule
            & 3 Epochs \\
            \midrule
            Datasets & BookCorpus \& English Wikipedia \\
            \greyrule
            Batch size & 256 \\
            \greyrule
            Initial learning rate & 5e-4 \\
            Learning rate schedule & linear decay with rewinds \\
            Learning rate rewinds & periodic every 0.5 epochs \\
            \greyrule
            Max sequence length & 512 \\
            Weight decay & 0.01 \\
            \greyrule
            \makecell{Knowledge Distillation\\(hardness, temperature)} & (1.0, 5.5) \\
            \greyrule
            Student model & prepared upstream teacher \\
            Teacher model & prepared upstream teacher \\
            \greyrule
            Pruning frequency & 4x per epoch \\
            \bottomrule
            \end{tabular}
        }
            \caption{Upstream pruning hyper-parameters.}
            \label{tab:hyperparams-UpstreamPruning}
\end{table}

\begin{table}
      \centering
        {\small 
            \begin{tabular}{l|c}
            \toprule
            & 8 Epochs \\
            \midrule
            Initial learning rate & 1.5e-4 \\
            Learning rate schedule & linear decay to 1.5e-6 \\
            \greyrule
            Batch size & \makecell{16 for SQuAD,\\ 32 for GLUE} \\
            \greyrule
            \makecell{Knowledge Distillation\\(hardness, temperature)} & (1.0, 5.5) \\
            \greyrule
            Teacher model & \bert \\
            \bottomrule
            \end{tabular}
        }
        \caption{Sparse-transfer learning hyper-parameters used to fine-tune upstream-pruned models at downstream tasks. These hyper-parameters are used to obtain results presented in Table \ref{tab:sparse-transfer}.}
    \label{tab:hyperparams-transfer}
\end{table}

\subsection{Downstream quantization}
\label{app:hyperparams-DownstreamQuantization}

We perform QAT on top of dense and 4-block pruned models on SQuAD v1.1 as shown in Table \ref{tab:downstream-block}. We quantize to 8 bits the embedding matrices, linear modules of all encoder units which includes matrices in their attention and feed forward layers, and the linear module of the output layer. Weights that were pruned are kept constant (zero) during quantization (sparsity mask preserved). Non-linear operations within the Softmax, LayerNorm and GeLU are not quantized. For each dense and 4-block pruned model in Table \ref{tab:downstream-block}, we perform a total of ten epochs training where the quantization observers are active for the first five and the remaining is fine-tuning. We do hyper-parameter search over the learning rates of 1e-4, 8e-5, 5e-5, 3e-5 and the distillation hardness of 0.9 and 1.0. We then pick the model with the best F1 score.

\subsection{Additional performance metrics}
Due to the space constraints, in the paper we report F1 score for SQuAD v1.1, matched accuracy for MNLI, and accuracy for QQP dataset. As all of our hyper-parameters for MNLI and QQP are exactly the same, we refer to these two datasets as GLUE. In Table \ref{tab:10_30_gradual_v2} we report the additional metrics too: exact match (EM) for SQuAD v1.1, mismatched accuracy for MNLI, and F1 score for QQP dataset. Tables \ref{tab:10_30_gradual_stdev} and \ref{tab:sparse-transfer-deviations} present standard deviations of the corresponding results in Tables \ref{tab:10_30_gradual}, \ref{tab:sparse-transfer} and \ref{tab:10_30_gradual_v2}. Finally, Table \ref{tab:downstream-block-additional} presents the exact-match metric for the corresponding results in Table \ref{tab:downstream-block}.

\begin{table}[htb!]
\setlength{\tabcolsep}{4pt}
    \centering
    {\small 
    \begin{tabular}{ccc|cc|c}
    \toprule 
    Task & \makecell{BERT\\\small{BASE}} & Sparsity & \makecell{Soft\\MvP} & \makecell{oBERT\\(ours)} & \makecell{oBERT\\(ours)}\\
    \midrule
      & Epochs && \multicolumn{2}{c|}{10 Epochs} & 30 Epochs\\                       
    \midrule
    \makecell{SQuAD\\EM} & 81.22 & \makecell{80\% \\ 90\% \\ 97\%} & \makecell{- \\ 76.60 \\ 72.70} &  \makecell{- \\ \textbf{80.76} \\ \textbf{76.14}} & \makecell{\textbf{82.08} \\ \textbf{81.12} \\ \textbf{78.11}}\\
    \midrule
    \makecell{MNLI\\mm-acc} & 85.06 & \makecell{80\% \\ 90\% \\ 97\%} & \makecell{- \\ 81.80 \\ 80.10} & \makecell{- \\ \textbf{83.58} \\ \textbf{80.67}} &  \makecell{\textbf{84.91} \\ \textbf{84.35} \\ \textbf{82.01}}\\
    \midrule
    \makecell{QQP\\F1} & 88.00 &  \makecell{80\% \\ 90\% \\ 97\%} & \makecell{- \\ 86.80 \\ 85.50} & \makecell{- \\ \textbf{87.69} \\ \textbf{87.05}} & \makecell{\textbf{88.63} \\ \textbf{88.30} \\ \textbf{87.66}}\\
    \bottomrule
    \end{tabular}
    }
    \caption{Additional evaluation metrics for results presented in Table \ref{tab:10_30_gradual}.}
    \label{tab:10_30_gradual_v2}
\end{table}

\begin{table}[htb!]
    \centering
    {\small 
    \begin{tabular}{ccc|c|c}
    \toprule 
    Task & \makecell{BERT\\\small{BASE}} & Sparsity & \makecell{Prune\\OFA} & \makecell{oBERT\\(ours)} \\
    \midrule
    \makecell{SQuAD \\ EM} & 81.42 & \makecell{90\% \\ 97\%} & \makecell{79.83 \\ -} & \makecell{\textbf{81.43} \\ 76.90} \\
    \midrule
    \makecell{MNLI \\ mm-acc} & 85.06 & \makecell{90\% \\ 97\%} & \makecell{82.43 \\ -} & \makecell{\textbf{83.78} \\ 81.13} \\
    \midrule
    \makecell{QQP \\ F1} & 88.00 & \makecell{90\% \\ 97\%} & \makecell{87.72 \\ -} & \makecell{\textbf{87.81} \\ 86.97} \\
    \bottomrule
    \end{tabular}
    }
    \caption{Additional evaluation metrics for results presented in Table \ref{tab:sparse-transfer}.}
    \label{tab:upstream-additional}
\end{table}

\begin{table}[htb!]
    \centering
    {\small
    \begin{tabular}{c|c|c|cc}
    \toprule
    Layers & Sparsity & Unstructured & 4-block & +QAT \\
    \midrule
    12 &  \makecell{0\% \\ 80\% \\90\%} & \makecell{82.71 \\ 82.08 \\ 81.12} & \makecell{82.71 \\ 81.46 \\ 80.14} & \makecell{81.99 \\ 80.57 \\ 78.84}\\
    \midrule
    6 & \makecell{0\% \\ 80\% \\90\%} & \makecell{81.17 \\ 81.15 \\ 79.16} & \makecell{81.17 \\ 79.55 \\ 77.65}  & \makecell{80.85 \\ 78.27 \\ 76.56}\\
    \midrule
    3 & \makecell{0\% \\ 80\% \\90\%}  & \makecell{76.62 \\ 75.62 \\ 73.61} & \makecell{76.62 \\ 74.07 \\ 71.36} & \makecell{76.06 \\ 72.70 \\ 70.00} \\
    \bottomrule
    \end{tabular}
    \caption{Additional evaluation metric (exact-match) for results presented in Table \ref{tab:downstream-block}.}
    \label{tab:downstream-block-additional}
    }
\end{table}

\begin{table}[htb!]
      \centering
         {\small 
             \begin{tabular}{cc|c}
                \toprule 
                Task & Sparsity & \makecell{oBERT\\ (ours)}\\
                \midrule
                  & Epochs & 30 Epochs\\
                \midrule
                \makecell{SQuAD\\ F1, EM} & \makecell{80\% \\ 90\% \\ 97\%} & \makecell{0.11, 0.03 \\ 0.13, 0.13 \\ 0.11, 0.17}\\
                \midrule
                \makecell{MNLI\\m, mm} & \makecell{80\% \\ 90\% \\ 97\%} & \makecell{0.14, 0.13 \\ 0.05, 0.04 \\ 0.35, 0.22}\\
                \midrule
                \makecell{QQP\\ acc, F1} &  \makecell{80\% \\ 90\% \\ 97\%} & \makecell{0.08, 0.08 \\ 0.04, 0.06 \\ 0.05, 0.08}\\
                \bottomrule
            \end{tabular}
         }
     \caption{Standard deviations for results presented in Tables \ref{tab:10_30_gradual} and \ref{tab:10_30_gradual_v2}.}
    \label{tab:10_30_gradual_stdev}
\end{table}

\begin{table}[htb!]
      \centering
            {\small 
                \begin{tabular}{cc|c}
                \toprule 
                Task & Sparsity & \makecell{oBERT \\(ours)} \\
                \midrule
                \makecell{SQuAD \\ F1, EM} & \makecell{90\% \\ 97\%} & \makecell{0.13, 0.13 \\ 0.03, 0.14} \\
                \midrule
                \makecell{MNLI \\ m, mm} & \makecell{90\% \\ 97\%} & \makecell{0.08, 0.24 \\ 0.17, 0.35} \\
                \midrule
                \makecell{QQP \\ acc, F1} & \makecell{90\% \\ 97\%} & \makecell{0.06, 0.07 \\ 0.09, 0.18} \\
                \bottomrule
                \end{tabular}
            }
    \caption{Standard deviations for results presented in Table \ref{tab:sparse-transfer} and \ref{tab:upstream-additional}.}
    \label{tab:sparse-transfer-deviations}
\end{table}

\subsection{Inference speedups and compression ratios of compressed models}
Details on the results shown in Figure \ref{fig:inference} are drawn from Table \ref{tab:deepsparseresults}. As shown in the results, not all compound compressed models yield improvements in inference or compression relative to retained model performance but those that do allow for massive improvements.

\begin{table*}[h!]
\setlength{\tabcolsep}{4pt}
    \centering
    {\small 
    \begin{tabular}{ccccccccc}
    \toprule 
    \makecell{Layers\\\phantom{x}} & \makecell{Sparsity\\(\%)} & \makecell{Compression\\Method} & \makecell{F1 score\\\phantom{x}} & \makecell{F1 recall\\(\%)} & \makecell{Throughput\\(items per sec.)} & \makecell{Speedup\\DeepSparse} & \makecell{Model size\\ (gzip MB)} & \makecell{Compression\\Ratio (w.r.t. gzip)} \\
    \midrule
    12 & 0 & none & 88.54 & 100.00 & \phantom{11}65.81 & \phantom{1}1.00 & 384.7 & \phantom{1}1.00 \\
    \greyrule
    12 & 80 & unstructured & 89.04 & 100.56 & \phantom{1}222.66 & \phantom{1}3.38 & 173.1 & \phantom{1}2.22 \\
    12 & 90 & unstructured & 88.31 & \phantom{1}99.74 & \phantom{1}292.40 & \phantom{1}4.44 & 140.1 & \phantom{1}2.75 \\
    12 & 80 & 4-block+QAT & 87.89 & \phantom{1}99.26 & \phantom{1}552.22 & \phantom{1}8.39 & \phantom{1}37.8 & 10.18\\
    \greyrule
    6 & 80 & unstructured & 88.20 & \phantom{1}99.62 & \phantom{1}419.68 & \phantom{1}6.38 & 128.3 & \phantom{1}3.00 \\
    6 & 90 & unstructured & 86.78 & \phantom{1}98.01 & \phantom{1}663.02 & 10.07 & 111.8 & \phantom{1}3.44 \\
    6 & 80 & 4-block+QAT & 86.10 & \phantom{1}97.24 & \phantom{1}989.54 & 15.04 & \phantom{1}26.2 & 14.70\\
    \greyrule
    3 & 80 & unstructured & 84.08 & \phantom{1}94.96 & \phantom{1}737.62 & 11.21 & 105.9 & \phantom{1}3.63 \\
    3 & 90 & unstructured & 82.50 & \phantom{1}93.18 & \phantom{1}974.00 & 14.80 & \phantom{1}97.7 & \phantom{1}3.94 \\
    3 & 80 & 4-block+QAT & 82.04 & \phantom{1}92.66 & 1892.27 & 28.75 & \phantom{1}20.3 & 18.92\\
    \bottomrule
    \end{tabular}
    }
    \caption{Compression effects on model size and inference speed, evaluated at batch size 32 with sequence length 128 on SQuAD v1.1 dataset. Evaluated at the c5.12xlarge AWS instance.}
    \label{tab:deepsparseresults}
\end{table*}

\subsection{Responsible NLP Research - Reproducibility Checklist}

In addition to many items from the ``Reproducibility Checklist'' which are already carefully addressed throughout the paper and Appendix sections, here we provide the remaining details to facilitate reproducibility of our results.

\subsubsection{Scientific Artifacts}
\noindent\textbf{Datasets.} Our experiments use existing and well established benchmarks for pre-training and fine-tuning of LLMs. Each dataset was used without any additional forms of modifications. Given that we did not modify any of the datasets, we did not inspect for personal, sensitive, or offensive content, nor did we perform any kind of anonymization. For pre-training, we make use of the Toronto Book Corpus (TBC)~\cite{Zhu2015AligningBA}~\footnote{https://huggingface.co/datasets/bookcorpus} and the wikipedia.20200501.en~\cite{wikidump}~\footnote{https://huggingface.co/datasets/wikipedia}. For fine-tuning we make use of SQuAD v1.1~\cite{Rajpurkar2016SQuAD1Q}~\footnote{https://huggingface.co/datasets/squad}, Quora Duplicate Question Dataset (QQP)~\cite{Shankar2017IdentifyingQQ}~\footnote{ https://huggingface.co/datasets/glue}, and Multi-Genre Natural Language Inference (MNLI)~\cite{N18-1101}~\footnote{ https://huggingface.co/datasets/glue} datasets. All these datasets are publicly available via HuggingFace datasets repository~\cite{hf-datasets}. The terms of usage and further details on each dataset can be found in their respective repositories.

\noindent\textbf{Models.} The model used as a starting point for all of our experiments is \bert, publicly available via HuggingFace Hub~\footnote{https://huggingface.co/bert-base-uncased}. All other models presented in this paper will be released in openly-available repositories along with their compression recipes, training metrics and hyper-parameters.

\subsubsection{Dataset Statistics}
Dataset statistics are detailed in Table \ref{tab:dataset_stat}.
\begin{table}[!htb]
      \centering
         {\small 
             \begin{tabular}{l|c|c}
                \toprule 
                Dataset & Train & Eval \\
                \midrule
                SQuAD (examples) &  87599 & 10570\\
                \midrule
                MNLI (examples) & 392702 & 19628 \\
                \midrule
                QQP (examples) &  363,846 & 40,430 \\
                \midrule
                Wikipedia (words) & 6078422  & - \\
                \midrule
                TBC (words) & 74004228 & - \\ 
                \bottomrule
            \end{tabular}
         }
     \caption{Statistics for training and evaluation datasets}
    \label{tab:dataset_stat}
\end{table}

\subsubsection{Computational Experiments}
\noindent\textbf{Upstream.} All upstream runs are in general computationally expensive due to the large batch sizes and huge datasets. In our experiments we make use of 4x A100 40GB NVIDIA GPUs. In this configuration, a single training epoch takes approximately 6 hours. Since the cost of such a large compute instance is high, these experiments were only run with a single seed and without major hyper-parameter exploration.

\noindent\textbf{Downstream.} Our downstream experiments make use of various different GPU cards that were at out disposal: 16GB V100, 11GB RTX 2080 Ti, and 24GB RTX 3090. Each training epoch takes approximately 30 minutes, and as a result the 30 epoch runs take approximately 15 hours. For these experiments, we report mean results of three runs with different random seeds.

\noindent\textbf{DeepSparse inference.} We pair our compressed models with DeepSparse~\cite{deepsparse} a publicly-available sparsity-aware CPU inference engine. This CPU runtime can leverage both structured and unstructured sparsity, and quantization to deliver high performance on commodity CPUs. We ran DeepSparse on a 24-core Intel AWS c5.12xlarge server with 24 cores, 96 vCPUs, 192 GB of RAM and an AVX-512 compatible instruction set. All models are exported using the standard ONNX\footnote{https://onnx.ai/} format.

\subsubsection{Computational Packages}
Our experiments build on publicly available libraries to ensure ease of reproduction and extensibility. All of our implementations, training and evaluation code are built on top of HuggingFace's Transformers \footnote{https://github.com/huggingface/transformers} and Datasets \footnote{https://github.com/huggingface/datasets} libraries, NeuralMagic's SparseML \footnote{https://github.com/neuralmagic/sparseml} library for model compression, and their DeepSparse \footnote{https://github.com/neuralmagic/deepsparse} engine for efficient inference on commodity CPUs.

\end{document}